\documentclass[journal]{IEEEtran}

\pdfoutput=1

\ifCLASSINFOpdf
  \usepackage[pdftex]{graphicx}
  \usepackage{subfigure}
  \usepackage{multirow}
  \usepackage{booktabs}
  \usepackage{amssymb}
  \usepackage{gensymb}
  \usepackage{color}
  \usepackage{amsmath,bm}

  \usepackage{diagbox}
\usepackage{subfigure}
\else
  \usepackage[dvips]{graphicx}
  \usepackage{subfigure}
  \usepackage{multirow}
  \usepackage{booktabs}
  \usepackage{amssymb}
  \usepackage{gensymb}
  \usepackage{color}
  \usepackage{amsmath,bm}

  \usepackage{diagbox}
\usepackage{subfigure}
\fi

\begin{document}

\title{Gated SwitchGAN for multi-domain facial image translation}

\renewcommand{\thefootnote}{\fnsymbol{footnote}}
\author{\IEEEauthorblockN{Xiaokang~Zhang\IEEEauthorrefmark{2},
		Yuanlue~Zhu\IEEEauthorrefmark{2},
		Wenting~Chen,
		Wenshuang~Liu, and
		Linlin~Shen\IEEEauthorrefmark{1}}
}

\markboth{IEEE TRANSACTIONS ON MULTIMEDIA}%
{Shell \MakeLowercase{\textit{et al.}}: Bare Demo of IEEEtran.cls for IEEE Journals}

\maketitle
\footnotetext[2]{The two authors contributed equally to the work.} 
\footnotetext[1]{Corresponding author: Prof. Linlin Shen, Tel: 86-0755-86935089, Fax: 86-0755-26534078, llshen@szu.edu.cn.} 
\footnotetext{The work was supported by the National Natural Science Foundation of China under grants no. 61672357, 91959108 and U1713214, and the Science and Technology Project of Guangdong Province under grant no. 2018A050501014.}
\footnotetext{X. Zhang, Y. Zhu, W. Chen, W. Liu and L. Shen are with the Computer Vision Institute, School of Computer Science \& Software Engineering, Guangdong Key Laboratory of Intelligent Information Processing, Shenzhen University, Shenzhen, 518060, China. 
}

\begin{abstract}
Recent studies on multi-domain facial image translation have achieved impressive results. The existing methods generally provide a discriminator with an auxiliary classifier to impose domain translation. However, these methods neglect important information regarding domain distribution matching. To solve this problem, we propose a switch generative adversarial network (SwitchGAN) with a more adaptive discriminator structure and a matched generator to perform delicate image translation among multiple domains. A feature-switching operation is proposed to achieve feature selection and fusion in our conditional modules. We demonstrate the effectiveness of our model.
Furthermore, we also introduce a new capability of our generator that represents attribute intensity control and extracts content information without tailored training. Experiments on the Morph, RaFD and CelebA databases visually and quantitatively show that our extended SwitchGAN (i.e., Gated SwitchGAN) can achieve better translation results than StarGAN, AttGAN and STGAN. The attribute classification accuracy achieved using the trained ResNet-18 model and the FID score obtained using the ImageNet pretrained Inception-v3 model also quantitatively demonstrate the superior performance of our models.
\end{abstract}

\IEEEpeerreviewmaketitle

\section{Introduction}

\IEEEPARstart{I}{mage} translation is the task of translating an image into one or several specific domains. 
The traditional approaches \cite{laffont2014transient,shih2013data,fergus2006removing,korshunova2017fast} addressed different tasks of image translation, e.g., changing the time of day or weather for outdoor images \cite{laffont2014transient,shih2013data}, removing camera shake from a single photograph \cite{fergus2006removing,korshunova2017fast} or face swapping. However, each of these tasks was tackled with a separate, special-purpose designed model, but a robust model capable of jointly achieving these functions was expected. 

Driven by the technique of generative adversarial networks (GANs) \cite{GANs,Pix2Pix,Pix2PixHD,CycleGAN,cGANs,ComboGAN,StarGAN,AttGAN,SEG-I2I,SwapGAN,FUNIT,StyleGAN,Image2StyleGAN,shi2020can}, image translation has witnessed various breakthroughs. The conditional GANs (cGANs) \cite{cGANs} provide conditional information to both the generator and discriminator to synthesize the desired image and become the base GAN model for image translation. When cGAN-based approaches \cite{cGANs,ACCE} generate images with a random vector $z$, they also introduce uncertainty for the task of image translation. Subsequently, Pix2Pix \cite{Pix2Pix} solves this problem in a supervised manner using cGANs. They propose a deterministic structure that learns the mapping from the source image, instead of the random vector $z$, to a target image. Wang et al. proposed an improved version of Pix2Pix \cite{Pix2PixHD} with feature matching loss for high-resolution image-to-image translation. However, their work requires paired samples, which poses an obstacle for training. For example, in age progression and regression \cite{ACCE,AgeCGAN}, it is nearly impossible to collect paired faces spanning a long range of years for every individual. To tackle this shortcoming, a cycle-consistent loss \cite{CycleGAN} and a quality-aware loss \cite{QGAN} are utilized in GANs to enforce the under-constrained translation and solve the problem of mapping the source domain to the target domain without paired samples. Attribute features have also been extracted in \cite{DFI,DFI2} for face translation. Moreover, some recent works \cite{lu2019face,lu2020global} focusing on face super-resolution, may be used to further improve the quality of facial image translation. However, all these translation methods aim to translate an image from one domain to another and are inefficient for a cross-domain translation.

A challenge with multi-domain image translation is how to achieve significant attribute translation. Some approaches have been proposed to address this problem. IcGANs \cite{ICGAN} encode an input image into a latent representation $z$ and attribute information $y$ by separate encoders and then apply changes to $y$ to generate the image with desired attributes. However, their synthesized images show low quality. Several approaches, i.e., ComboGAN \cite{ComboGAN}, StarGAN \cite{StarGAN}, AttGAN \cite{AttGAN} and STGAN \cite{STGAN}, have achieved impressive results. In ComboGAN, the CycleGAN \cite{CycleGAN} structure is expanded to a multi-domain translation system. Although the approach achieves high-quality translation, the system requires a massive computational cost. StarGAN, AttGAN and STGAN train a conditional network via an auxiliary classification domain loss \cite{ACGAN} and reconstruction loss \cite{CycleGAN,AttGAN}. These three methods neglect the importance of domain distribution matching during multiple domain image translation. Although an auxiliary domain classification loss is imposed on the discriminator and generator to achieve multi-domain translation, it is inefficient to distinguish real or fake input in the discriminator without the corresponding domain information. In other words, an auxiliary classifier can easily distinguish the target class of a synthesized image and converge via slight image manipulation. However, it is far from photo-realistic. These problems become more obvious for facial age progression and regression (merely translating the wrinkle while neglecting the face shape and hair) or facial expression translation (merely translating the mouth while neglecting the eyes). Moreover, controlling image generation by label channels and the commonly shared generator limit the performance of image translation. 

Recently, style-guided methods \cite{StyleGAN, Image2StyleGAN} have shown great potential for reference based image generation. For the image translation task, although AdaIN \cite{AdaIN} is incorporated into the content and style decomposition structure for multi-modal translation, MUNIT \cite{MUNIT} cannot handle significant appearance variations between the two domains. FUNIT \cite{FUNIT} and StarGAN v2 \cite{starganV2} introduce the information of the style image into the decoder to achieve multi-domain image translation. However, such style-guided methods mainly focus on the diversity of image translation and the similarity between the reference and generated image, which is different from the target of multi-domain facial attribute translation. For this purpose, the translation of facial attributes is usually controlled by target attribute labels. Moreover, since the classification output of the discriminator is used for adversarial learning in \cite{FUNIT, starganV2}, the performance and stability are limited. While style-guided models benefit from multi-class classifiers based on reconstruction and adversarial losses, they still neglect the importance of domain information, which limits their performance.

To address the problems above, we propose a novel generative adversarial network, namely, SwitchGAN, to achieve multiple domain facial image translation with more remarkable and adaptive attribute translation effects. The network consists of two conditional modules: a generator with an attribute encoder switch and a discriminator with an attribute adversary switch. Different from the previous works \cite{StarGAN,AttGAN,STGAN} that concatenate the input image or feature maps with their corresponding label, our label is regarded as a switch of different attribute features in the networks. Our generator achieves the desired translation by the attribute encoder switch. Similarly, an attribute adversary switch is also used in our discriminator to provide a stronger conditional constraint and a more adaptable structure for multi-domain image translation. The primary work of our SwitchGAN \cite{SwitchGAN} was published in ICME. The translation accuracy and visual results reported in different models demonstrate the superior performance of our SwitchGAN. 

Despite the superior performance of our SwitchGAN, it could not simultaneously control the attribute intensity during image translation, which is very useful for real applications. ExprGAN \cite{ExprGAN} controls the intensity of expression translation by an intensity controller module, which maximize the mutual information \cite{InfoGAN} between the generated image and the expression code. Fader networks \cite{FaderNetworks} encode the input image into information unrelated to the attribute by imposing an adversarial constraint and apply an attribute label to manipulate the intensity of the output. Other methods \cite{DFI,DFI2} extract attribute features for intensity control. In this paper, we significantly extend our previous work by incorporating a gate module and a conditional feature matching loss into the SwitchGAN. The gate module, with a common shared encoder and an attribute encoder switch, is proposed to control the intensity of face translation. Moreover, with the gate module, the performance of SwitchGAN is improved. The conditional feature matching loss, which is designed specifically for our discriminator structure, forces our model to produce domain-specified natural statistics at multiple scales and improves the performance of our generator. The extended SwitchGAN, namely Gated SwitchGAN, is capable of synthesizing face images with different attribute intensities, by setting different values of the attribute encoder switch output. Compared to our previous work, much more intensive experiments have been conducted and popular image quality indices such as the FID score \cite{FID, InceptionV3} have also been reported to evaluate the quality of synthesized face images.

Overall, our major contributions are listed as follows:
\begin{itemize}
	\item We present a novel framework for face translation. In this framework, an effective discriminator with an attribute adversary switch and a generator with an attribute encoder switch are proposed to achieve image translations with controllable intensities among multiple domains. 
	\item Based on SwitchGAN, a gate module is proposed to control the intensity of attribute translation and further improve the performance of the SwitchGAN models. A conditional feature matching loss is proposed to help the generator to synthesize high-quality images.
	\item We visually and quantitatively evaluate the performance of the proposed SwitchGAN and Gated SwitchGAN. Both translation accuracy and FID score suggest that our framework can achieve better translation results than state-of-the-art models.
\end{itemize}

\section{SwitchGAN: Switch Generative Adversarial Network}
\begin{figure*}[t]
  \centering
  \includegraphics[width=0.83\linewidth]{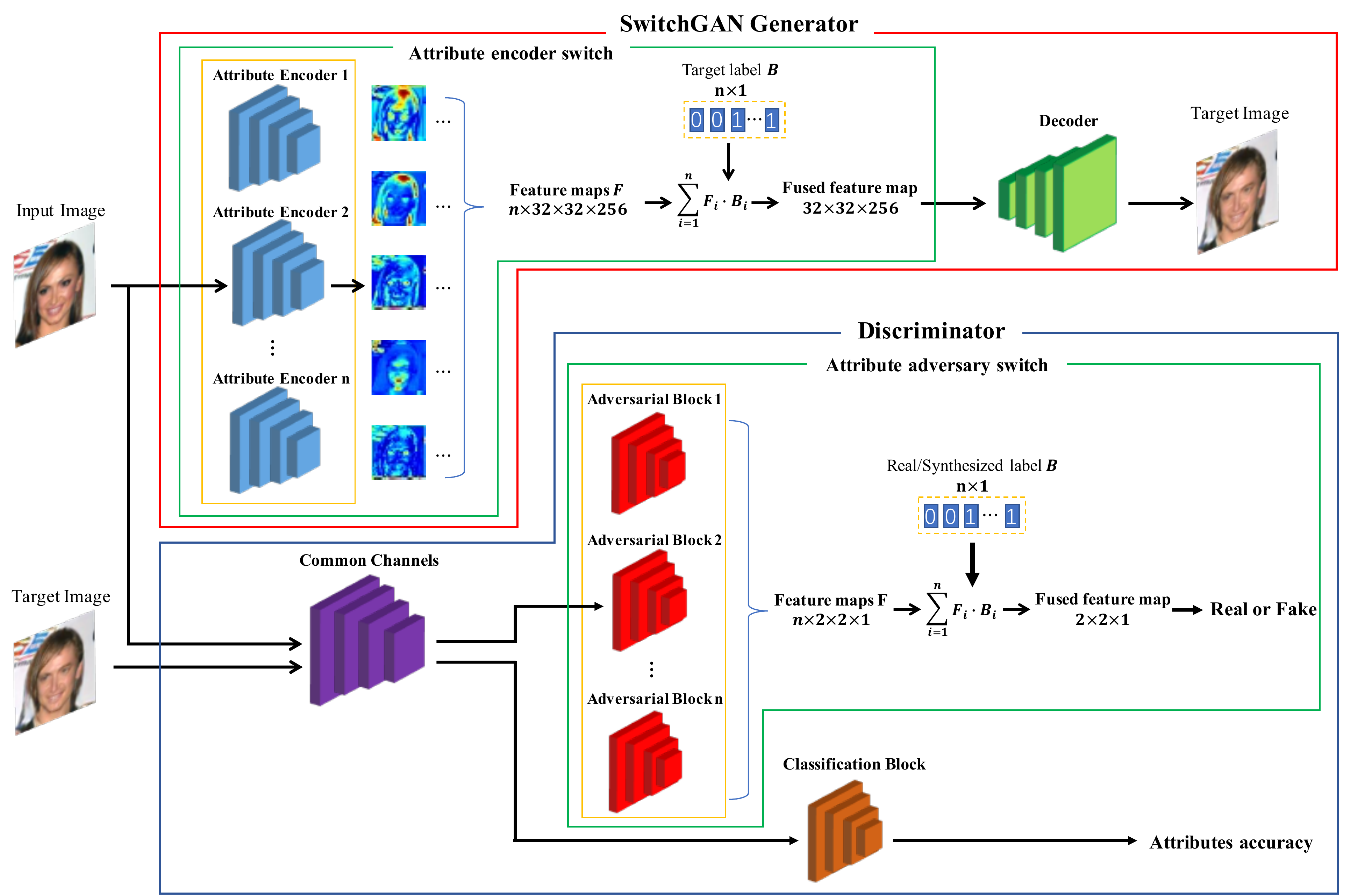}
  \caption{Architecture of the proposed SwitchGAN. The $128 \times 128 \times 3$ real face represents the input image, and the non-zero component in the $n \times 1$ binary target label represents the attributes to be translated for the face.}
  \label{SwitchGAN_architecture}
\end{figure*}

\begin{figure*}
	\centering
	\includegraphics[width=0.83\linewidth]{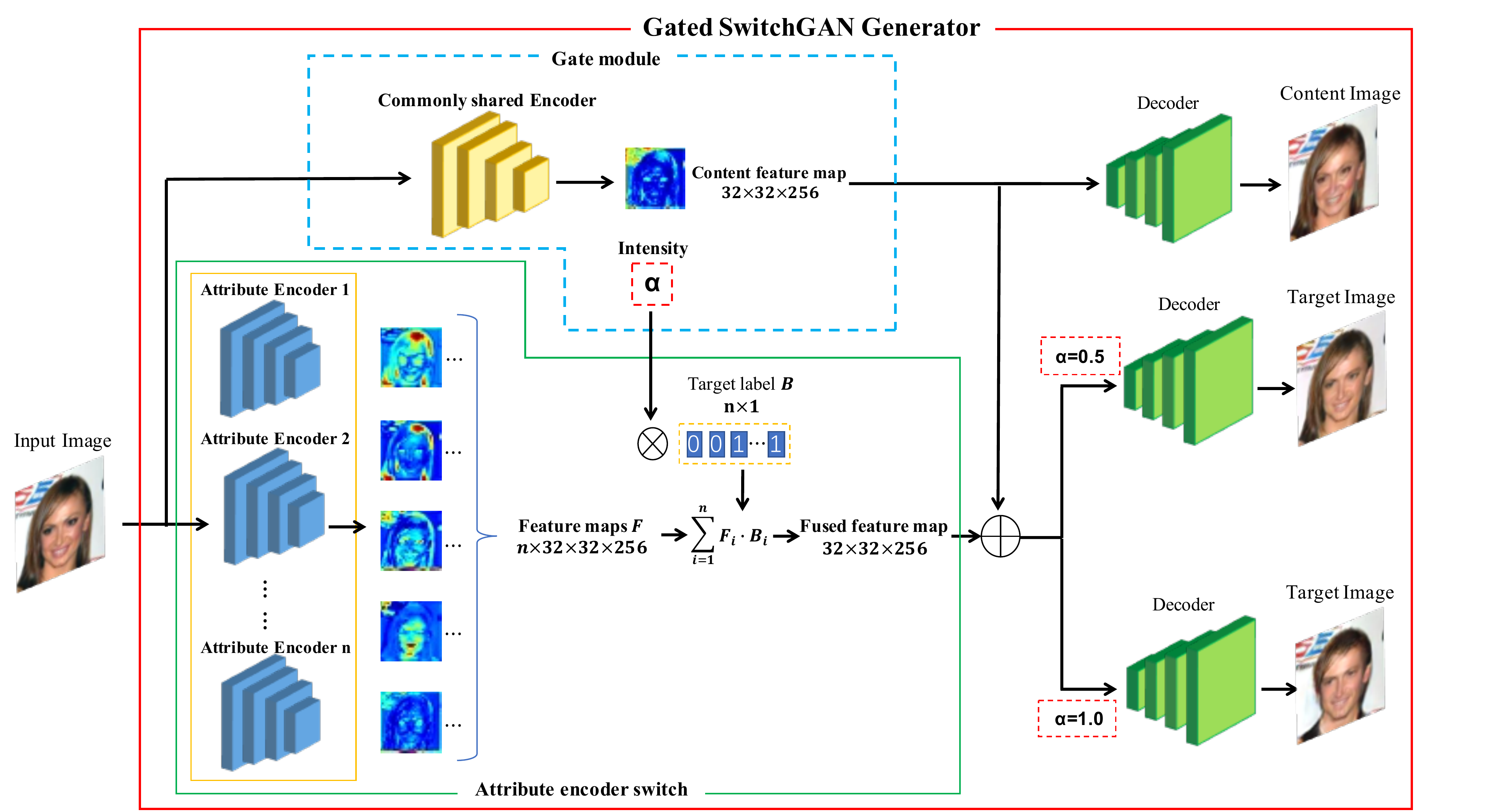}
	\caption{Architecture of the proposed generator for Gated SwitchGAN. Different from the SwitchGAN generator, a commonly shared encoder is added to encode the content information. Note that we fix $\alpha$ = 1 ($\alpha_{i}$ = 1, i = 1...n) during the whole training stage, while for the testing stage, we can continuously change $\alpha$ in [0, 1] ($\alpha_{i} \in [0,1]$, i = 1...n) to control the intensity of attributes.}
	\label{GatedSwitchGAN_architecture}
\end{figure*}

In this section, we introduce the novel SwitchGAN and Gated SwitchGAN in detail. We propose a SwitchGAN that utilizes a generator with an attribute encoder switch and a discriminator with an attribute adversary switch to achieve multi-domain facial image translation. A feature-switching operation is applied to achieve feature selection and fusion in both of our conditional switch modules. In our SwitchGAN, the label vector is used as a switch to decide the desired attribute features in the generator and discriminator, whereas it is often concatenated to the feature map in the previous approaches. The Gated SwitchGAN significantly extends the above work by incorporating a gate module into the SwitchGAN. The gate module, which imposes a commonly shared encoder in our SwitchGAN generator, exhibits the flexibility to control the intensity of desired attributes and further improves the performance of SwitchGAN. Different from SwitchGAN where the label vector is merely used as a switch to decide the desired attributes features, Gated SwitchGAN also uses the attribute encoder switch as a gate of the attribute intensity during translation. Fig. \ref{SwitchGAN_architecture} illustrates the training process of the proposed SwitchGAN approach. Fig. \ref{GatedSwitchGAN_architecture} illustrates the generator structure of the Gated SwitchGAN.

\subsection{Architecture}
{\bf Generator.}
As shown in Fig. \ref{SwitchGAN_architecture}, our generator consists of an attributes encoder switch module with parallel attribute encoder branches $E_{i},i \in (1,n)$, and a single decoder module $M$, where both of the modules are trained at the same time. Our generator takes an image $x$ and a target label vector $c_{trg}$ as input. Image $x$ is encoded by the encoder branches first, and then the encoded feature maps $F$ are multiplied with the target label vector $c_{trg}$ and summed up to generate fused features $F_{fusion}$. Both the original label vector $c_{org}$ and the target label ${c_{trg}}$ can be represented as follows:
\begin{equation}
c=[B_{1},B_{2},\cdots,B_{n}]
\end{equation}
where $B_{i}$ is the binary label of the $i$-th attribute. The feature maps $F$ input to the feature-switching operator can also be represented as follows:
\begin{equation}
F=[F_{1},F_{2},\cdots,F_{n}]
\end{equation}
For the generator, $F_{i}$ represents the feature map encoded by the $i$-th encoder $E_{i}$. Then, the feature-switching operation can be formulated as follows:
\begin{equation}
F_{fusion}=\sum_{i=1}^{n} (F_{i} \cdot B_{i})
\end{equation}
The binary label works as a switch, and only the feature maps corresponding to the non-zero component of label vector pass to the following process. As a result, our generator automatically selects the desired attribute information for multiple domain facial image translation.

It is important to learn the multi-domain joint distribution via a single generator. In one-to-one domain image translation, both the shared common latent space \cite{UNIT} and the content and style decomposition \cite{MUNIT} provide special structures to learn such a joint distribution. Inspired by their works, we provide a latent space for the fused features to achieve joint distribution, and then a single decoder successfully learns the mapping from the joint distribution to the desired output. Consequently, our flexible generator is capable of adaptive switching among one-to-many or many-to-many image translation tasks and provides a better interpretability than the previous works. Note that the input label vector must have at least one non-zero value.

For SwitchGAN, the function of the generator can be formulated as follows:
\begin{equation}\label{func4}
M(\sum_{i=1}^{n} (E_{i}(x) \cdot B_{i}))
\end{equation}
To control the intensity of attribute translation and improve the performance, we introduce a gate module for the SwitchGAN generator. The gate module consists of a commonly shared encoder $E_{m}$ and the intensity of attributes $\alpha$. The shared encoder, with the same structure of the attribute encoder, is added to the SwitchGAN generator. As shown in Fig. \ref{GatedSwitchGAN_architecture}, the output of the attribute encoder switch is fused with the output of the shared encoder. Then, a similar single decoder successfully learns the mapping from the joint distribution to the desired output. The generator function can be formulated as follows:
\begin{equation}\label{func5}
M(E_{m}(x) + \sum_{i=1}^{n} (E_{i}(x) \cdot B_{i}))
\end{equation}
Both the SwitchGAN and Gated SwitchGAN share the same training process, while for the testing process, we introduce a vector $\alpha$ to enable the model to control the intensity of translation, which can be represented as follows:
\begin{equation}
\alpha=[\alpha_{1},\alpha_{2},\cdots,\alpha_{n}]
\end{equation}
where $\alpha_{i}$ $\in$ [0,1] represents the intensity of the $i$-th attribute. Therefore, the value of each attribute intensity can continuously vary from 0 to 1, and the generator function during testing process can be formulated as follows:
\begin{equation}\label{func7}
M(E_{m}(x) + \sum_{i=1}^{n} (E_{i}(x) \cdot \alpha_{i} \cdot B_{i}))
\end{equation}
Since only the attribute encoders corresponding to the non-zero label go through the attribute encoder switch, the commonly shared encoder should learn to balance the features among different attribute encoders. In other words, the shared encoder is indirectly forced to encode the content information and the intermediate state of all attributes. As shown in Fig. \ref{GatedSwitchGAN_architecture}, the content feature map can be decoded to generate a content image, which retains basic semantic content of our input image. As a result, the attribute encoder switch is more focused on the attribute information and improves the ability of translation. When we impose different values on the switch (i.e., $\alpha$ = 0.5 or 1), the output of the generator can show different intensities of attribute translation and simultaneously preserve the content information. The switch works as a gate of attributes. The higher the intensity value we impose, the clearer the attribute appearance it shows. 
Note that the motivation of the gate module is different from the cycle-consistent network~\cite{CycleGAN, ComboGAN}, i.e., the gate module is mainly designed to extract the common information and intermediate content.
We demonstrate in some experiments that owing to the shared encoder, the Gated SwitchGAN can visualize the variant intensity of translation and the content information of the input image. Please refer to the supplementary material for more details of the generator.

{\bf Discriminator.}
We utilize the same structure of the discriminator as that of the SwitchGAN models. As shown in Fig. \ref{GatedSwitchGAN_architecture}, our discriminator extracts shallow features by the common channels $D_{m}$, which is followed by a classification branch $D_{cls}$ and an attribute adversary switch $D_{adv}$. The former classifies the attributes of the input image, while the latter distinguishes whether the input image is real or fake.
 The discriminator function for the classification can be formulated as follows:
\begin{equation}
D_{cls} (D_{m}(x))
\end{equation}

Innovatively, the attribute adversary switch $D_{adv}$ consists of parallel adversarial blocks $D_{adv_{i}}$, where the feature-switching operation that is similar to the generator is utilized for the outputs of adversarial blocks. The features are multiplied with the label vector and summed to generate the final output, which can be formulated as follows:
\begin{equation}
\sum_{i=1}^{n} (D_{adv_{i}} (D_{m}(x)) \cdot B_{i})
\end{equation}
where $D_{adv_{i}}$ represents the specific adversary block for the input with the $i$-th attributes. Similar to the generator, only the features corresponding to the non-zero value of the label vector can go through the switch module.

The key idea of our discriminator is that the adversarial blocks discriminate the fake image as true and converge the model only if the synthesized image is indistinguishable from the real samples corresponding to the specific attributes. In contrast, the discriminator in former methods \cite{StarGAN,AttGAN} always directly discriminates the fake image as true and converges the model once the synthesized image is indistinguishable from the samples from database. We realize the importance of the neglected domain distribution information. Therefore, we provide conditional information for the discriminator to distinguish whether the input image is from real or synthesized samples. It is the reason why our model synthesizes images with more apparent attributes and higher quality.

Consequently, we provide a more effective conditional discriminator for multi-domain image translation tasks. The details of the discriminator network are illustrated in the supplementary material.

\subsection{Objective Function}
Given a facial image $x$ with an attribute $c_{org}$ and a target label vector $c_{trg}$, the generator aims to synthesize a face image with desired attribute $c_{trg}$. 

{\bf Adversarial loss.}
To conduct the training process of attribute translation, a conditional adversarial constraint is required. The adversarial function is denoted as follows:
\begin{equation}\label{func10}
\begin{array}{ll}
loss_{adv}&=\mathop{min}\limits_{G} \mathop{max}\limits_{D} E_{x,c_{org}}[D(x,c_{org})]\\
&- {E_{x,c_{trg}}[D(G(x,c_{trg}), c_{trg})]} \\
\end{array}
\end{equation}

It can be separated as $loss_{adv_{D}}$ and $loss_{adv_{G}}$ and ensures that the generator translates the image into the desired output.

{\bf Reconstruction loss.} 
A cycle-consistent loss \cite{CycleGAN} and a self-reconstruction loss \cite{AttGAN} are applied to preserve the content information of the output image. The cycle-consistent loss is formulated as follows:
\begin{equation}\label{func11}
loss_{cyc}=E_{x,c_{org},c_{trg}}[||x-G(G(x,c_{trg}),c_{org})||_1]
\end{equation}
The self-reconstruction loss is formulated as follows:
\begin{equation}\label{func12}
loss_{self}=E_{x,c_{org}}[||x-G(x,c_{org})||_1]
\end{equation}

{\bf Classification loss.} 
A classification loss \cite{StarGAN,ACGAN} is applied to the generator to promote the attribute classification accuracy.
The classification loss optimizes $D$ by real samples and optimizes $G$ by synthesized samples. By minimizing this objective, it forces the generated image to fall into the target domain. The classification loss function is denoted as follows:
\begin{eqnarray}\label{func13and14}
loss_{c_{org}} = &E_{x,c_{org}}[-logD(c_{org}|x)] \\
loss_{c_{trg}} = &E_{x,c_{trg}}[-logD(c_{trg}|G(x,c_{trg}))]
\end{eqnarray}

{\bf Conditional feature matching loss.}
As feature matching loss \cite{Pix2PixHD,SPADE} can encourage the model to produce natural statistics at multiple scales, it has been popularly utilized to optimize the generator. Let $D_{i}(x,c_{org})$ denote the feature map extracted from the $i$-$th$ down-sampling layer of the discriminator D for input x and its corresponding label $c_{org}$, and ${\bar{D}}_{i}(x,c_{org})$ denote the global average pooling result of $D_{i}(x,c_{org})$. The feature matching loss can be formulated as follows: 
\begin{equation}\label{func15}
\begin{array}{ll}                
loss_{fm}&=E_{x,c_{org},c_{trg}}\sum\limits_{i=1}^{T}[|| \bar{D}_{i}(x,c_{org})\\
&-{ \bar{D}_{i}(G(G(x,c_{trg}),c_{org}),c_{org})||_1] } \\
\end{array}
\end{equation}
where T denotes the amount of the selected feature maps in D. 

Deriving benefit from the attribute adversary switch $D_{adv}$, the discriminator can converge the model more smoothly. Furthermore, we can utilize the domain-speciﬁed feature maps extracted from parallel adversary blocks $D_{adv_{i}}$ of $D_{adv}$ to optimize the generator. Specifically, we construct the domain-speciﬁed feature extractor named $D^{f}_{adv_{i}}$, by removing the last (prediction) layer from $D_{adv_{i}}$. Then, we use $D^{f}_{adv_{i}}$ to extract the domain-speciﬁed information and implement the feature-switching operation as follows:
\begin{equation}
D^{f}_{adv}(x, c)=\sum_{i=1}^{n} (D^{f}_{adv_{i}} (D_{m}(x)) \cdot B_{i})
\end{equation}
With the fused domain-specified feature map obtained from the feature-switching operation, we finally propose a conditional feature matching loss as follows:
\begin{equation}\label{func16}
\begin{array}{ll}                
loss_{cfm}&=E_{x,c_{org},c_{trg}}[|| D^{f}_{adv}(x,c_{org})\\
&-{ D^{f}_{adv}(G(G(x,c_{trg}),c_{org}),c_{org})||_1] } \\
\end{array}
\end{equation}

{\bf Domain-aware gradient penalty.}
The original GANs are powerful generative models but suffer from training instability. To stabilize the training process, a gradient penalty \cite{WGAN} is utilized in our network. Considering the input of the label vector in our discriminator structure, the loss function with the gradient penalty is modified as follows:
\begin{equation}\label{func17}
\begin{array}{ll}
loss_{adv_{D}}&=E_{\tilde{x} \sim P_{g} } [D(\tilde{x},c_{trg})]-E_{ x \sim P_{r} } [D(x,c_{org})] \\
&+{ 10 \cdot E_{\hat{x} \sim P_{\hat{x}}}[(||\bigtriangledown _{\hat{x}} D(\hat{x},c_{trg})||_{2}-1)^{2}] }  \\
\end{array}
\end{equation}
where $P_{\hat{x}}$ in the third part is a distribution randomly sampled from the real distribution $P_{x}$ and synthesized distribution $P_{\tilde{x}}$. The loss function of $loss_{adv_{G}}$ is modified as follows:
\begin{equation}\label{func18}
loss_{adv_{G}} = E_{x,c_{trg}}[-D(G(x,c_{trg}), c_{trg})]
\end{equation}

{\bf Full objective.}
As a result, to generate a plausible image with target attributes, we arrive at the objective function as follows:
\begin{equation}\label{func19}
\begin{array}{ll}
loss_{G}&=loss_{adv_{G}} + \lambda _{cyc}\cdot loss_{cyc} + \lambda _{self}\cdot loss_{self}  \\
&+~{ \lambda _{c}\cdot loss_{c_{trg}} + \lambda _{cfm}\cdot loss_{cfm}} \\
\end{array}
\end{equation}
\begin{eqnarray}\label{func20}
loss_{D}&=loss_{adv_{D}} + \lambda _{c}\cdot loss_{c_{org}} 
\end{eqnarray}
where $\lambda _{cyc}$, $\lambda _{self}$, $\lambda _{c}$ and $\lambda _{cfm}$ are hyperparameters to control the importance of each loss.

\section{Experiments}
In this section, we visually and quantitatively evaluate the performance of our SwitchGAN and Gated SwitchGAN in different tasks, such as age progression and regression, facial expression translation, and facial attributes translation.

We compare our method with state-of-the-art methods: StarGAN \cite{StarGAN}, AttGAN \cite{AttGAN} and STGAN \cite{STGAN}, all of which learn mappings between multiple domains. We first compare our approaches with several unsupervised multi-domain image-to-image translation baselines. Then, we compare the attribute interpolation performance of different methods. Finally, we evaluate the effectiveness of the proposed modules.

\subsection{Dataset}
{\bf Morph} \cite{Morph} is the largest publicly available longitudinal face database. Its Album 2 contains 55,134 images of 13,000 individuals with age labels ranging from 16 to 77 years. We separate the images into a training set with 50,020 images and a test set with 4,925 images. The images are separated into five groups with ages 11-20, 21-30, 31-40, 41-50, and 50+. 

{\bf The Radboud Faces Database} (RaFD) \cite{RaFD} is a face database with 8,040 pictures of 63 individuals displaying 8 emotional expressions. We only use the 4,824 frontal images in the database. Six individuals are randomly chosen as the test set. The images are separated into eight facial expression groups. 

{\bf CelebA} \cite{CelebA} is a large-scale face attributes dataset. It contains 202,599 face images, each with 40 attribute annotations. The dataset provides a standard list for dataset separation: 162,770 images for training, 19,868 images for validation and 19,961 images for testing. We use the given training set and validation set as our training set and the remaining 19,961 images as the test set.

\begin{figure*}
	\centering
	\includegraphics[width=0.69\linewidth]{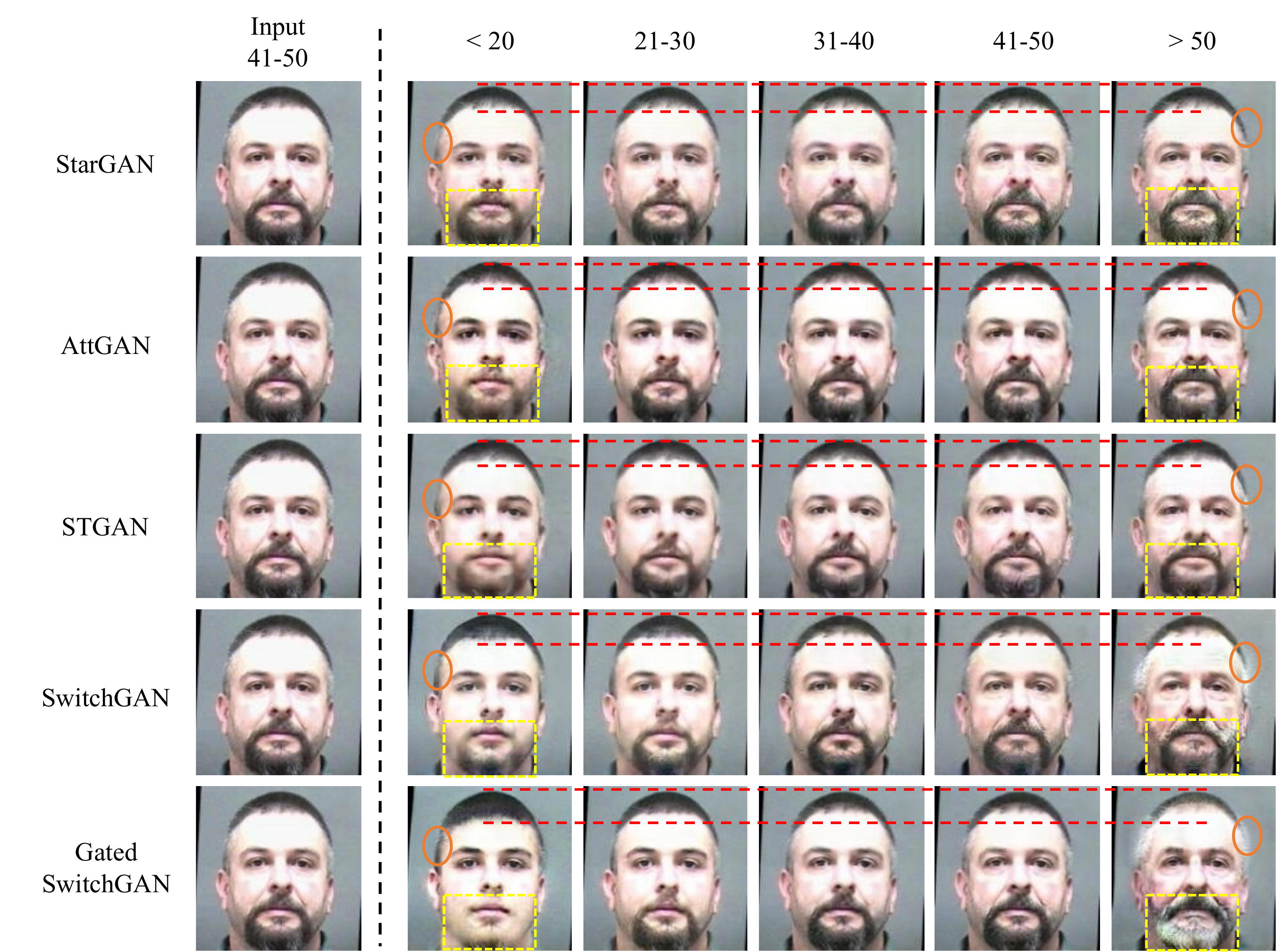}
	\caption{Results of age progression and regression on Morph. The first column represents the input image, while the remaining columns show various age groups translated by different models.}
	\label{morph_comparison}
\end{figure*}

\subsection{Training}
All SwitchGAN models are trained through the strategy of Adam grad with $\beta_1=0.5$ and $\beta _2=0.999$. We take 300,000 steps of training for the Gated SwitchGAN models in Morph and CelebA, and 120,000 steps of training for the Gated SwitchGAN in RaFD. The parameters are set as $\lambda _{cyc}=$ $\lambda _{self}=5$ and $\lambda _{c}=$ $\lambda _{cfm}=1$ for the Gated SwitchGAN models training during one-to-many tasks. We set $\lambda _{cyc}=$ $\lambda _{self}=10$, $\lambda _{c}=2$ and $\lambda _{cfm}=1$ for the experiment of the Gated SwitchGAN in CelebA \cite{CelebA}. The initial learning rates for discriminator and generator are 0.0001, which linearly decays to 0 at the second half steps. The batch size for all datasets are 16. The Morph database images are detected by the MTCNN \cite{MTCNN} and then aligned and cropped to $128 \times 128$. For the RaFD and CelebA databases, the images are cropped to $256 \times 256$, where the faces are centred, and resized to $128 \times 128$. The same data augmentation that flips the images horizontally with a probability of 0.5 is applied to the datasets. We use the function in (\ref{func17}) and the hinge version \cite{GeometricGAN} of GAN loss for training. For each training image, a randomly generated target label vector is used together with the image and associated attribute label vectors as input to the model for training. The generator is updated once when the discriminator is updated five times.

\subsection{Evaluation metrics}
If the trained generator shows a good performance, the synthesized images should be photo-realistic and follow the same distribution as the real samples. The attribute accuracy and Frechet Inception Distance (FID) score \cite{FID,InceptionV3} have been popularly used to evaluate the quality of synthesized images. We also adopt them to qualitatively evaluate the quality of our results.

{\bf The translation accuracy}, classified by a trained classifier, has been commonly utilized to evaluate image translation tasks. We employ it to compare the performances of different models. We utilize ResNet-18 \cite{ResNet} (a stable and parameter-insensitive model to report convincing results) and select the model with the best performance in the test set (MAE of 3.42 for age in Morph, accuracy of 99.54\% for facial expression in RaFD, and average accuracy of 93.76\% for five selected attributes in CelebA) for attribute classification. The synthesized images on the test set are tested by the classifiers. For example, when evaluating the model in Morph, each of the 4,925 test images is translated into the five age groups 11-20, 21-30, 31-40, 41-50, and 50+. As a result, 24,625 ($4,925 \times 5$) synthesized images are tested by the trained age classifier to assess the translation performance.

{\bf The FID score} is shown to correlate well with human judgement of visual quality by measuring the similarity between two datasets of images. It calculates the Wasserstein-2 distance between the generated images and the real images in the feature space of an Inception-v3 network. A lower FID score indicates that the distribution of generated images is more similar to that of real target images. Following the steps in \cite{FID}, a feature vector is first extracted by a pretrained network \cite{FID,InceptionV3} for each attribute. We then compute FID between every generated image group and the target ground truth images and report the mean FID.

\begin{table}
	\centering
	\caption{Quantitative comparison for different methods using the translation accuracy and FID score. For translation accuracy, higher is better. For FID score, lower is better.}
	\begin{tabular}{cccc}
	    \toprule[1pt]
		Database & Method & Accuracy$\uparrow$ & FID score$\downarrow$ \\
		\midrule
		Morph & StarGAN & 63.42\% & 18.31\\
		& AttGAN & 54.76\% & 17.37\\
		& STGAN & 70.06\% & 18.41\\
		& SwitchGAN & 76.94\%{\footnotemark[1]} & 17.45\\
		& Gated SwitchGAN & {\bf 85.74\%} & {\bf 15.3}\\
		\midrule
		RaFD & StarGAN & 92.53\% & 20.33\\
		& AttGAN  & 81.68\% & 29.21\\
		& STGAN  & 87.30\% & 28.92\\
		& SwitchGAN & 93.34\%{\footnotemark[2]} & 33.38\\
		& Gated SwitchGAN & {\bf 98.32\%} & {\bf 17.4}\\
		\bottomrule[1pt]
	\end{tabular}
	\label{baseline_comparison_1}
\end{table}

\footnotetext[1]{Different from the preprocessing procedures in our prior work \cite{SwitchGAN}, face detection and alignment based on MTCNN \cite{MTCNN}  are applied in this paper; we believe that face landmark-based alignment is a more reasonable and widely used procedure for preprocessing in face-related applications.}
\footnotetext[2]{We trained a new advanced facial expression classifier, which improved the accuracy of the previous model (used in \cite{SwitchGAN}) from 98.38\% to 99.54\%.}

\begin{table*}
	\centering
	\caption{Quantitative results of different methods on CelebA using translation accuracy and the FID score. For translation accuracy, higher is better. For FID score, lower is better.}
	\begin{tabular}{cccccccc}
		\toprule[1pt]
		Measures & Method & Black hair colour & Blond hair colour & Gender & Moustache & Age & Average \\
		\midrule
		Accuracy$\uparrow$ & StarGAN & 98.68\% & 92.50\% & 95.01\% &65.44\% & 94.52\% & 89.23\% \\
		& AttGAN & 71.35\% & 79.86\% & 90.52\% & 55.92\% & 93.57\% & 78.24\%  \\
		& STGAN & 85.28\% & 93.00\% & {\bf 97.71\%} & 85.15\% & 93.90\% & 91.01\% \\
		& SwitchGAN & 98.29\% & 86.94\% & 94.51\% & 83.37\% & 92.98\% & 91.22\%  \\
		& Gated SwitchGAN & {\bf 99.03}\% & {\bf 96.69\%} & 97.27\% & {\bf 86.15\%} & {\bf 96.84\%} & {\bf 95.20\%} \\
		\midrule
		FID Score$\downarrow$ & StarGAN & 6.3 & 18.7 & 51.27 & {\bf 59.09} & 20.12 & 31.09 \\
		& AttGAN & 6.16 & {\bf 18.31} & 48.75 & 60.76 & 19.26 & 30.65 \\
		& STGAN & {\bf 5.42} & 19.04 & 57.22 & 65.04 & 19.95 & 33.33\\
		& SwitchGAN & 9.31 & 24.37 & 42.69 & 66.95 & 21.95 & 33.05 \\
		& Gated SwitchGAN & 5.8 & 18.56 & {\bf 36.34} & 64.36 & {\bf 18.91} & {\bf 28.79} \\
		\bottomrule[1pt]
	\end{tabular}
	\label{baseline_comparison_2}
\end{table*}

\subsection{Quantitative comparison}
We calculate the translation accuracy and FID score for different baselines. The comparison is shown in Table \ref{baseline_comparison_1} and Table \ref{baseline_comparison_2}. 

{\bf Morph.}
An age group classifier (ResNet-18) trained using the training set of Morph is used to classify the images generated by different GAN models. If the synthesized face of a target group is correctly classified by the classifier, we decide such a translation is a successful one. Table \ref{baseline_comparison_1} shows the classification results for different models. While the classifier achieves 63.42\%, 54.76\% and 70.06\% accuracy on the images generated by StarGAN, AttGAN and STGAN, respectively, it achieves higher accuracy on the images generated by the SwitchGAN (76.94\%) and Gated SwitchGAN models (85.74\%). A pretrained Inception-v3 model is utilized to compute the FID score. The FID scores are reported in Table \ref{baseline_comparison_1}. Our Gated SwitchGAN model achieves the best FID score over the previous methods.

{\bf RaFD.} As shown in Table \ref{baseline_comparison_1}, the classifier achieves 92.53\%, 81.68\% and 87.30\% accuracy on the images generated by StarGAN, AttGAN and STGAN, respectively, while our SwitchGAN and Gated SwitchGAN models achieve higher accuracy, i.e., 93.34\% and 98.32\%, respectively. A pretrained Inception-v3 model is again utilized to compute the FID score. Our Gated SwitchGAN model achieves the best FID score among different methods.

\begin{figure*}
	\centering
	\includegraphics[width=0.77\linewidth]{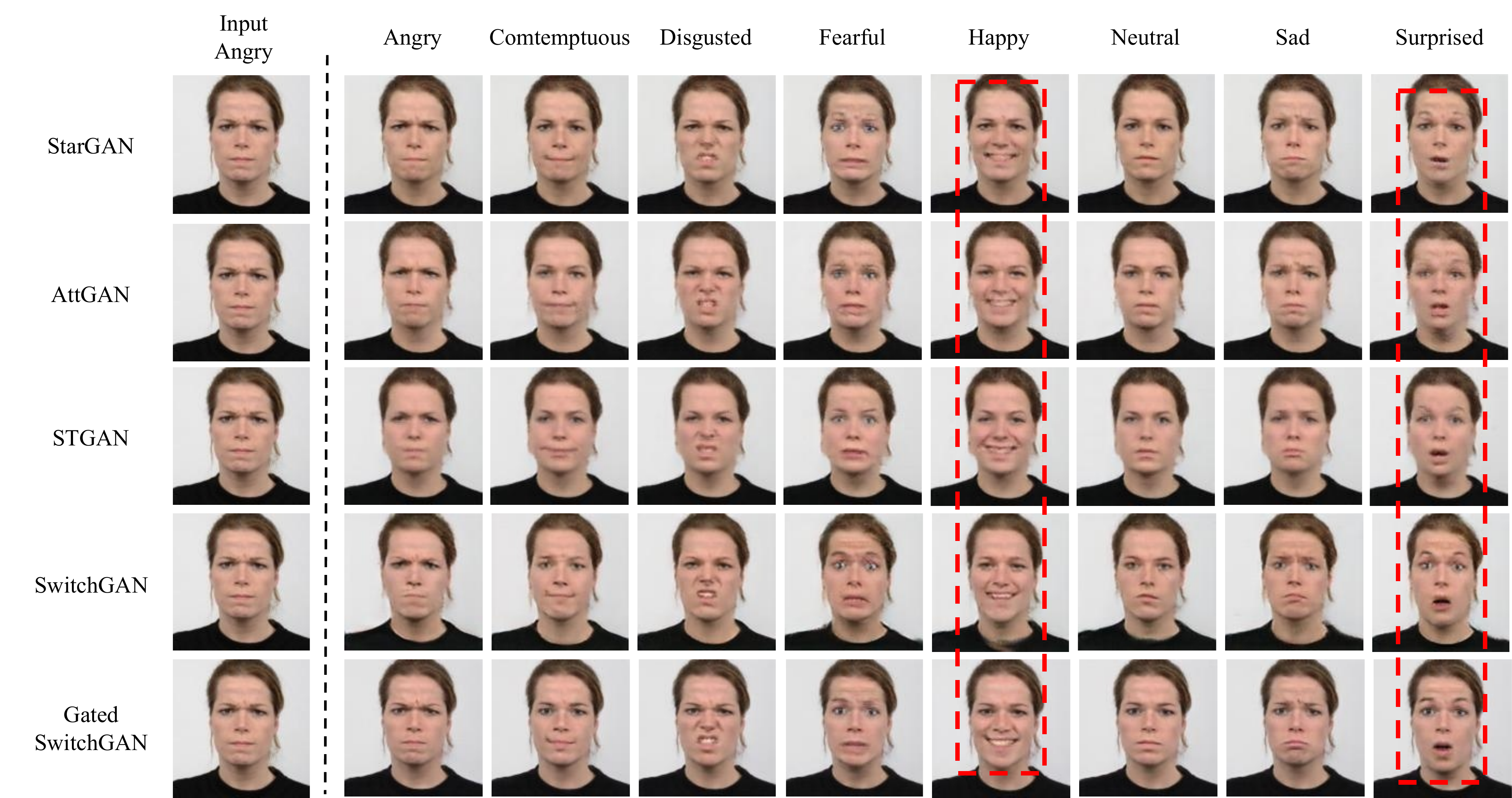}
	\caption{Results of facial expression translation on RaFD. The first column represents the input image, while the remaining columns show the facial expressions translated by different models.}
	\label{rafd_comparison}
\end{figure*}

\begin{figure*}
	\centering
	\includegraphics[width=0.77\linewidth]{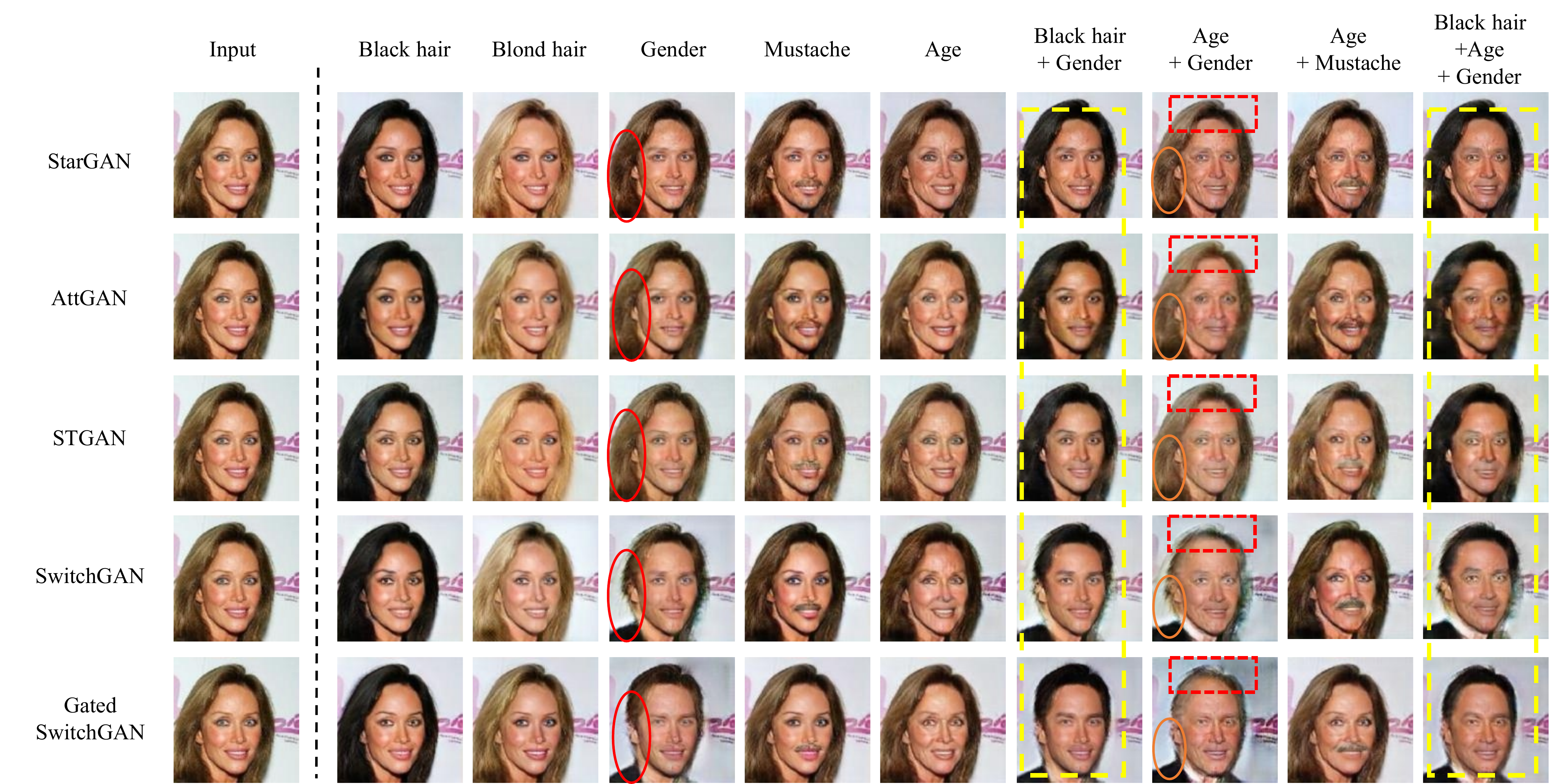}
	\caption{Results of multiple attribute translation on CelebA. The first column represents the input image, while the remaining columns show the synthesized attributes translated by different models.}
	\label{celeba_comparison}
\end{figure*}

{\bf CelebA.}
Table \ref{baseline_comparison_2} shows the classification results for different models. The classifiers achieve 89.23\%, 78.24\% and 91.01\% average accuracy on the images generated by StarGAN, AttGAN and STGAN, respectively, while our SwitchGAN and Gated SwitchGAN achieve higher accuracy, i.e., 91.22\% and 95.20\%, respectively. A pretrained Inception-v3 model is utilized to compute the FID scores, which are reported in Table \ref{baseline_comparison_2}. Our Gated SwitchGAN achieves the best average FID score, which demonstrates the superior ability of our Gated SwitchGAN model.

\subsection{Qualitative comparison}
Our SwitchGAN and Gated SwitchGAN models are mainly compared with StarGAN, AttGAN and STGAN for age progression and regression, facial expression translation, and multiple facial attributes translation.

\begin{figure*}
	\centering
	\includegraphics[width=0.78\linewidth]{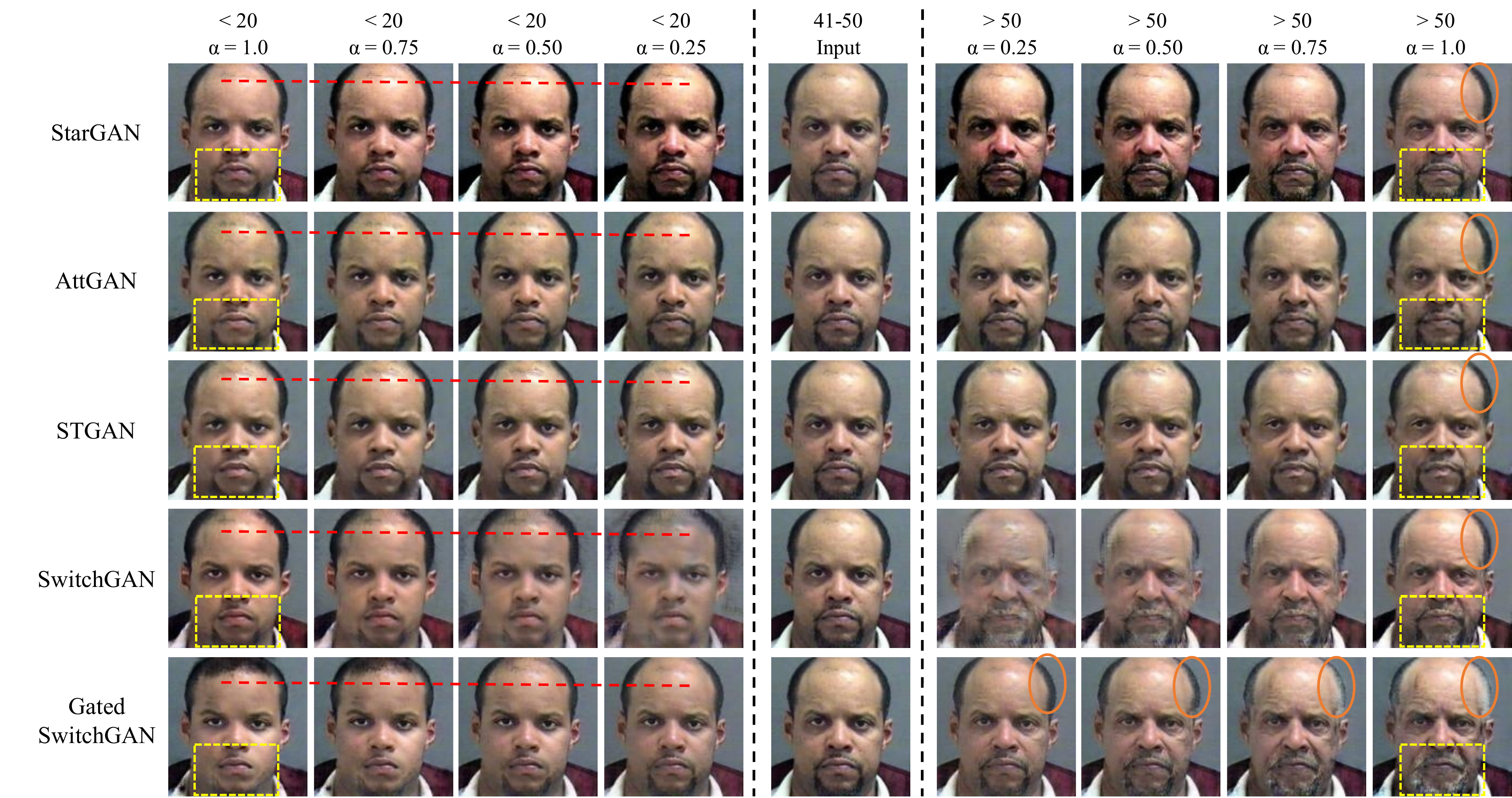}
	\caption{The translation results of Gated SwitchGAN on Morph, with different intensities. The input is translated to the age groups of younger than 20 years and older than 50 years with intensities of 0.25, 0.50, 0.75 and 1. 
	Note that we set all of the elements $\alpha_{i}$ of $\alpha$ to the same value here, e.g., $\alpha$ = 1 means $\alpha_{1}$ = $\alpha_{2}$ = ... = $\alpha_{i}$ = ... = $\alpha_{n}$ = 1.}
	\label{Morph_intensity}
\end{figure*}

\begin{figure*}
	\centering
	\includegraphics[width=0.78\linewidth]{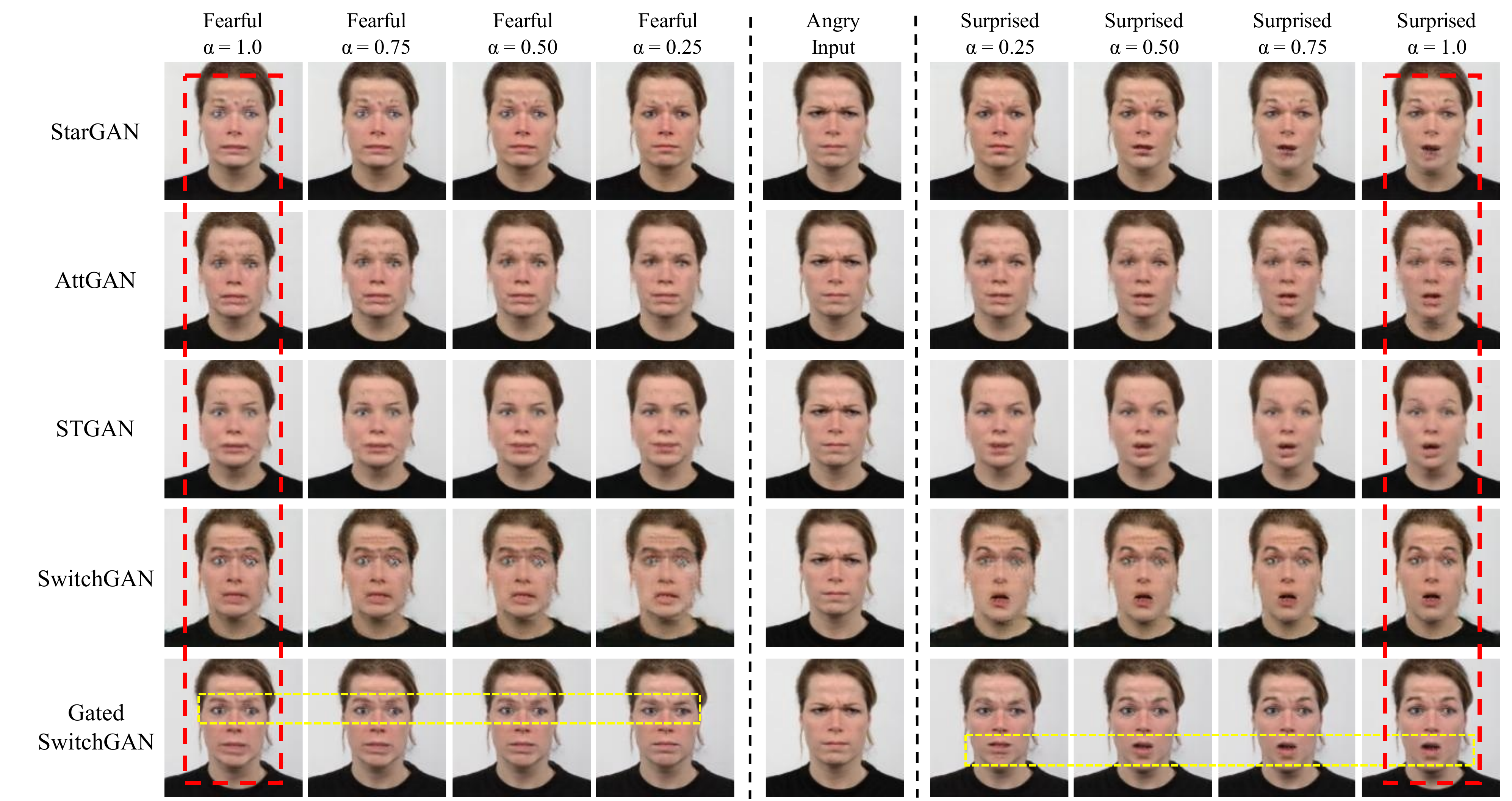}
	\caption{The translation results of Gated SwitchGAN on RaFD, with different intensities. The input is translated to fearful and surprised expressions with intensities of 0.25, 0.50, 0.75 and 1. 
	Note that we set all of the elements $\alpha_{i}$ of $\alpha$ to the same value here, e.g., $\alpha$ = 1 means $\alpha_{1}$ = $\alpha_{2}$ = ... = $\alpha_{i}$ = ... = $\alpha_{n}$ = 1.}
	\label{RaFD_intensity}
\end{figure*}

\begin{figure*}
	\centering
	\includegraphics[width=0.78\linewidth]{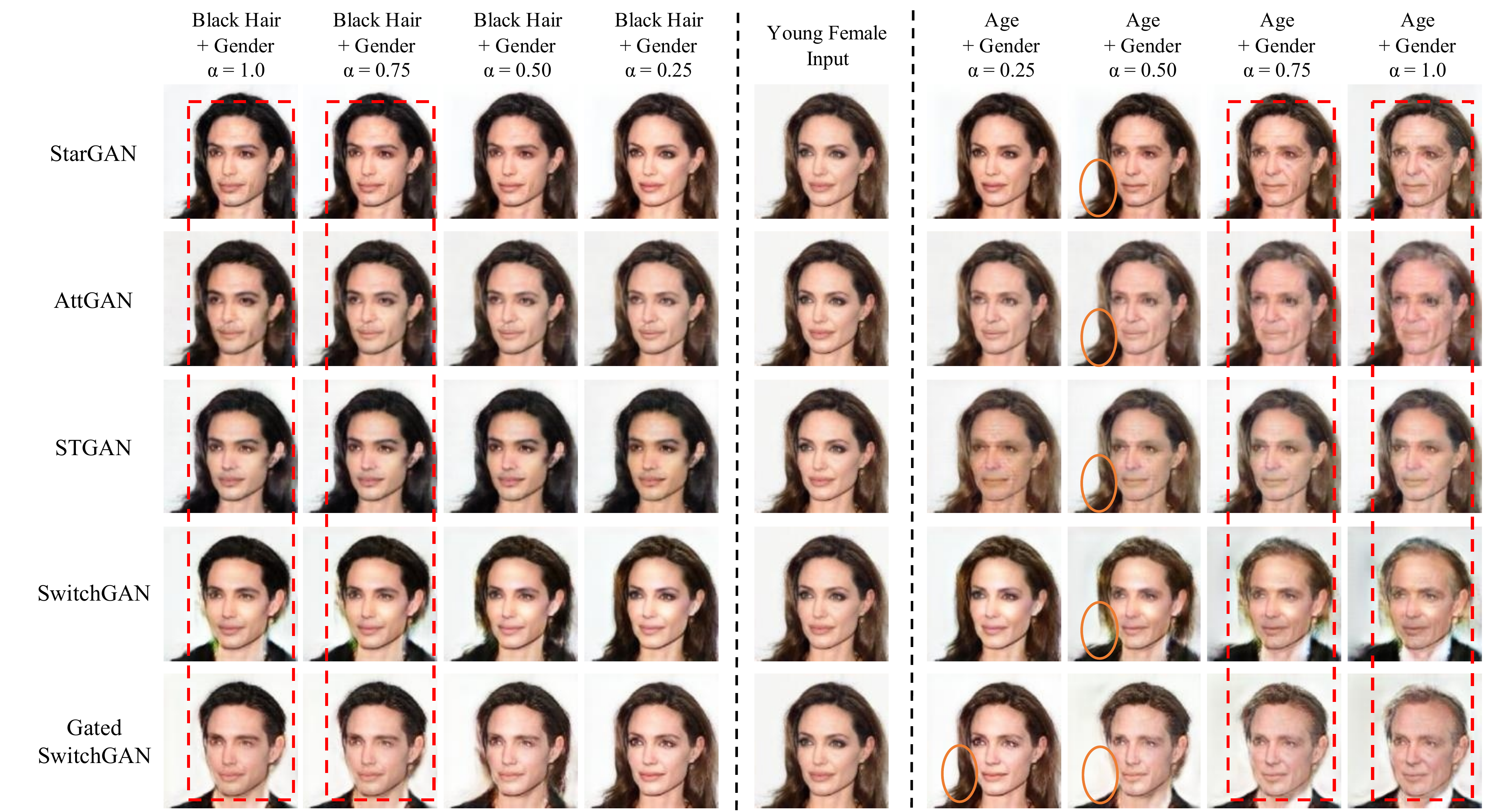}
	\caption{The translation results of Gated SwitchGAN on CelebA, with different intensities. The input is translated into an old man and a young man with black hair with intensities of 0.25, 0.50, 0.75 and 1. 
	Note that we set all of the elements $\alpha_{i}$ of $\alpha$ to the same value here, e.g., $\alpha$ = 1 means $\alpha_{1}$ = $\alpha_{2}$ = ... = $\alpha_{i}$ = ... = $\alpha_{n}$ = 1.}
	\label{CelebA_intensity}
\end{figure*}

{\bf Morph.}
Age progression and regression aims to translate a facial image into the desired age group by transforming wrinkles, hairlines, hair colours, etc. Fig. \ref{morph_comparison} shows the results of different models. In Fig. \ref{morph_comparison}, the example faces of a 41-year-old man translated by StarGAN, AttGAN, STGAN, SwitchGAN, and Gated SwitchGAN are visually compared. All the methods achieve reasonable age group translation. However, StarGAN, AttGAN and STGAN mainly adapt the wrinkle translation. The faces associated with ages of more than 50 years that are translated by our models present higher hair lines, much more white hair and beards to make them look older. The faces associated with ages less than 20 years that are translated by StarGAN, AttGAN and STGAN do not look young, while our SwitchGAN models regrow the black hair and shave the beard to make them look younger. Moreover, compared with SwitchGAN, the Gated SwitchGAN shows more significant changes in the beard and hair. The results clearly justify the usefulness of the SwitchGAN models.

{\bf RaFD.}
The facial expression translation aims to translate a facial expression into another. Fig. \ref{rafd_comparison} shows the results of different models. In Fig. \ref{rafd_comparison}, the faces of different expressions translated by StarGAN, AttGAN, STGAN, SwitchGAN and Gated SwitchGAN are visually compared. As shown in the figure, only the mouth is appropriately changed for the surprised face translated by StarGAN, AttGAN and STGAN, while the eyes still look unsurprised. Our SwitchGAN models change both the eyes and mouth to make them look surprised. Compared to SwitchGAN, Gated SwitchGAN shows more appealing changes are made to the mouth and eyes. Similar translation can be found for the happy expression, i.e., the mouth generated by Gated SwitchGAN is more realistic.

{\bf CelebA.}
The attributes of black hair, blond hair, gender, moustache, and age are selected for experiments. The first two attributes aim to translate a facial image into an output with black hair or blond hair. The remaining three attributes aim to translate a facial image into an output with the opposite attributes, e.g., translating male into female, or female into male. Fig. \ref{celeba_comparison} shows the results of different models. In Fig. \ref{celeba_comparison}, the example faces of a young woman with brown hair translated by StarGAN, AttGAN, STGAN, our SwitchGAN and Gated SwitchGAN are visually compared. The last three columns represent the joint translation of different attributes. As shown in the fourth column of Fig. \ref{celeba_comparison}, the young lady's hair remains long after the translation performed by StarGAN, AttGAN and STGAN, while the SwitchGAN models shorten the lady's hair to give it a more distinctively male appearance.
 As can be seen in the images of the 7th column, when the attributes of gender and age are translated at the same time, the images translated by StarGAN, AttGAN and STGAN merely manipulate the texture information of the input. The images translated by our SwitchGAN models shorten the hair and recede the hairline of the young lady to make her appear as if she were an old man. Moreover, as can be seen in images of the 6th and the last columns, the images translated by Gated SwitchGAN show less artefacts and more apparent results than SwitchGAN.

\begin{figure*}
	\centering
	\includegraphics[width=0.80\linewidth]{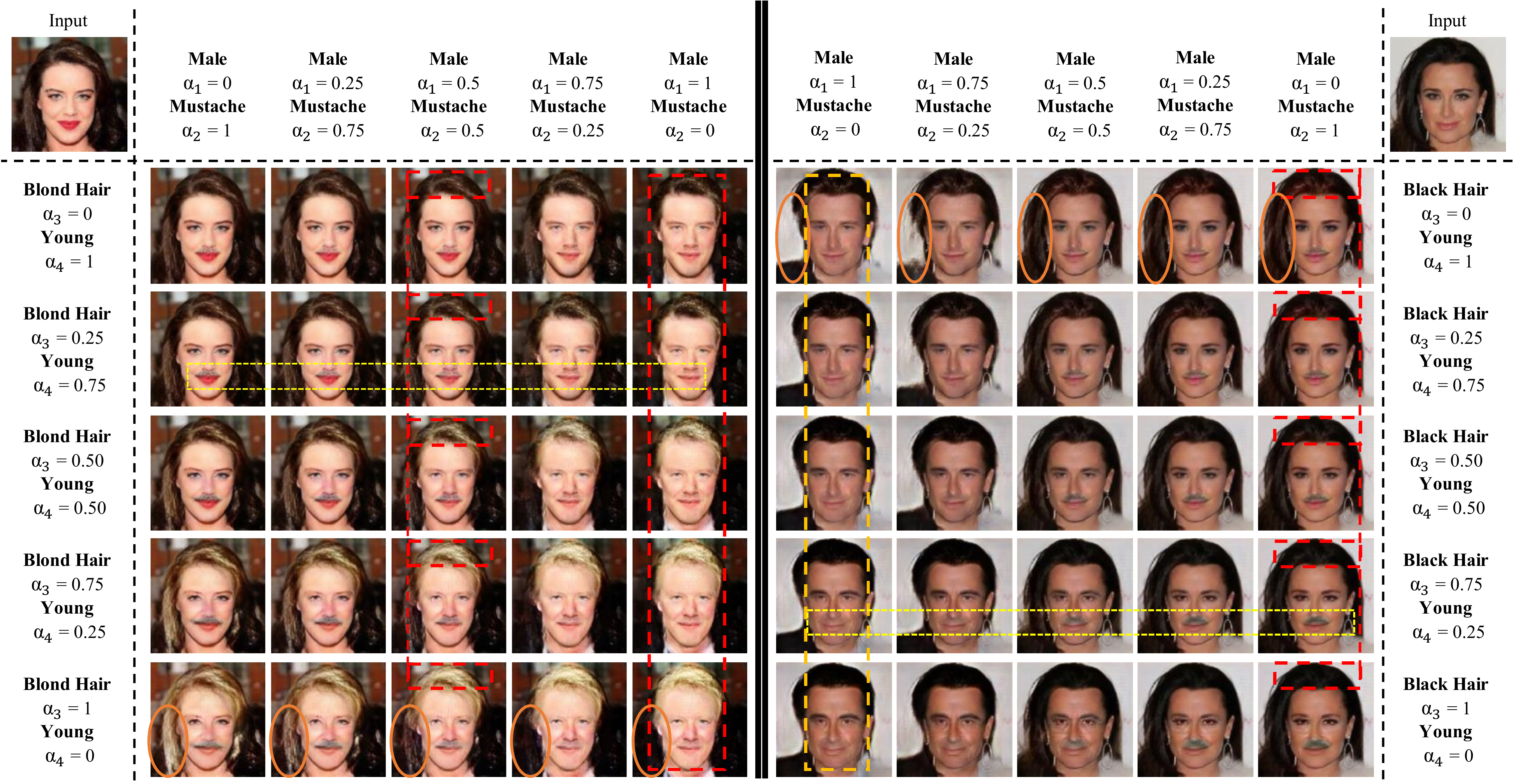}
	\caption{The translation results of Gated SwitchGAN on CelebA, with separate intensities for different attributes. While the faces in the row show the translation results obtained by changing the intensities of {\bf Male} ($\alpha_{1}$) and {\bf Mustache} ($\alpha_{2}$), the faces in the column present the results obtained by changing the intensities of {\bf Hair Colour} ($\alpha_{3}$) and {\bf Age} ($\alpha_{4}$).
	}
	\label{CelebA_intensity_cross}
\end{figure*}

\begin{figure*}
	\centering
	\includegraphics[width=0.78\linewidth]{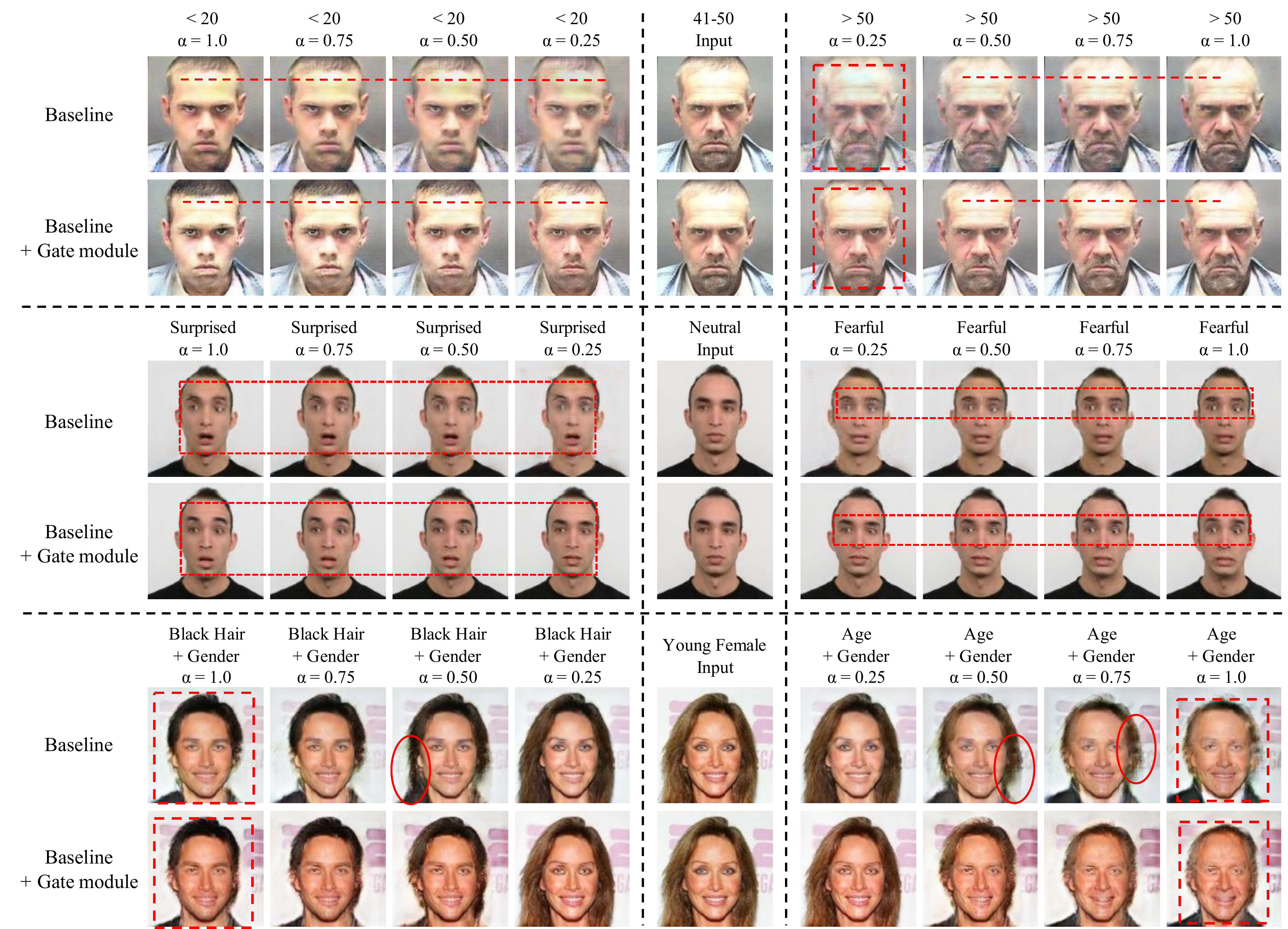}
	\caption{Qualitative comparison for the ablation study of the gate module on Morph, RaFD and CelebA datasets. On Morph, the input is translated to the age groups of younger than 20 years and older than 50 years. On RaFD, the input is translated to fearful and surprised expressions. On CelebA, the input is translated into an old man and a young man with black hair. All of these inputs are translated with intensities of 0.25, 0.50, 0.75 and 1.}
	\label{ablation_study_for_gate}
\end{figure*}

\begin{table*}[t]
    \centering
    \caption{Quantitative comparison for the ablation study using translation accuracy and FID score. For translation accuracy (Acc), higher is better. For FID scores, lower is better.}
    \begin{tabular}{c c c c c c c c c c c}
        \toprule[1pt]
        \multirow{2}{*}{Method} & \multirow{2}{*}{$L_{self}$} & \multirow{2}{*}{Gate} & \multirow{2}{*}{$L_{cfm}$} & \multirow{2}{*}{$L_{fm}$} & \multicolumn{2}{c}{Morph} & \multicolumn{2}{c}{RaFD} & \multicolumn{2}{c}{CelebA} \\
        \cmidrule{6-7} \cmidrule{8-9} \cmidrule{10-11} 
        
        & & & & & FID$\downarrow$ & Acc$\uparrow$ & FID$\downarrow$ & Acc$\uparrow$ & FID$\downarrow$ & Acc$\uparrow$ \\
            
        \midrule
        SwitchGAN & & & & & 17.45 & 76.94\% & 33.38 & 93.34\% & 33.05 & 91.22\%  \\
        \midrule
        (1) Baseline & $\checkmark$ & & & & 15.62 & 76.97\% & 27.45 & 95.23\% & 30.64 & 92.01\% \\
        \midrule
        (2) +Gate & $\checkmark$ & $\checkmark$ & &  & 15.35 & 83.38\% & 19.14 & 97.22\% & 29.29 & 93.00\% \\
        \midrule
        (3) +Gate+$L_{fm}$ & $\checkmark$ & $\checkmark$ &  &  $\checkmark$ & 16.41 & 79.34\% & 20.55 & 97.37\% & 29.41 & 94.38\% \\
        \midrule
        (4) +Gate+$L_{fm}$+$L_{cfm}$ & $\checkmark$ & $\checkmark$ & $\checkmark$ & $\checkmark$  & 16.05 & 82.66\% & 23.92 & 96.24\% & 29.88 & 92.53\% \\
        \midrule
        \textbf{(5) +Gate+$\bm{L_{cfm}}$} & $\checkmark$ & $\checkmark$ & $\checkmark$  & & \textbf{15.3} &\textbf{85.74\%} & \textbf{17.4} & \textbf{98.32\%} & \textbf{28.79} & \textbf{95.20\%} \\
        \bottomrule[1pt]
    \end{tabular}
    \label{ablation_study}
\end{table*}

\begin{figure*}
	\centering
	\includegraphics[width=0.96\linewidth]{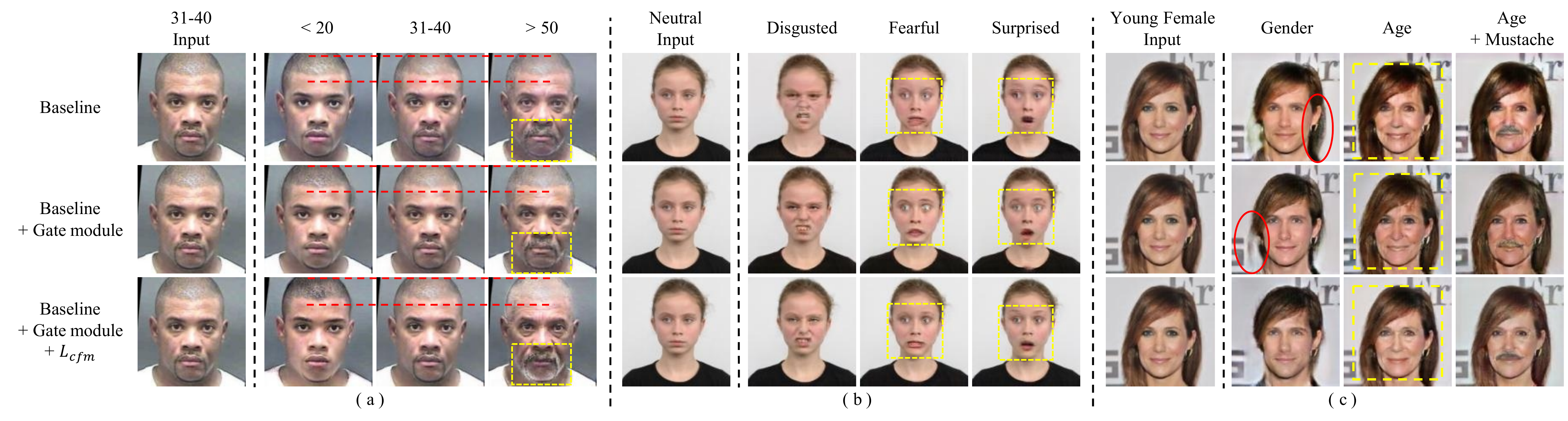}
	\caption{Qualitative comparison for the ablation study of $L_{cfm}$ on Morph (a), RaFD (b) and CelebA (c) datasets.}
	\label{ablation_study_for_cfm}
\end{figure*}

\subsection{Attribute intensity control.}
Given an input image, let $\alpha$ denote the weight value of target attributes $c_{trg}$, such that we can continuously control the attribute intensity of the output appearance by the function mentioned in Equation \ref{func7}. 

{\bf Morph}. The age translation with different intensities, i.e., $\alpha$ = 0.25, 0.5, 0.75 and 1, are shown in Fig. \ref{Morph_intensity}, where the face of a man is translated into the age groups of younger than 20 years and older than 50 years by different methods. When the value of the attribute intensity is continuously changed, the images translated by StarGAN, AttGAN, and STGAN merely adapt the wrinkle translation. For the images translated by SwitchGAN, if a low value of $\alpha$ is given, artefacts and vacuity appear in the images due to the loss of encoded information in the latent space. The images translated by Gated SwitchGAN represent different age characteristics when different intensity values are imposed. For example, when the face of the input man is translated to the age group of less than 20 years old, the higher the intensity we impose, the more the black hair and the less the beard that appear in the output. When the input face is translated to the age group of more than 50 years old, the higher the intensity we impose, the more the beard, the whiter the hair and the higher the hair line that appear in the output.

{\bf RaFD}. The facial expression translation with different intensities, i.e., $\alpha$ = 0.25, 0.5, 0.75 and 1, are shown in Fig. \ref{RaFD_intensity}, where an angry face is translated into fearful and surprised expressions by different methods. When the value of the attribute intensity changes continuously, the images translated by StarGAN, AttGAN, and STGAN remain stable and obvious artefacts can be observed. For the images translated by SwitchGAN, when the value of $\alpha$ is small, artefacts appear in the images due to the loss of encoded information in the latent space. The images translated by Gated SwitchGAN are smoothly controlled by the intensity value and considerably fewer artefacts can be observed. For example, when the angry face is translated to the surprised expression, the higher the intensity we impose, the wider the mouth and eyes open. Similar results can also be found for the translation of the fearful facial expression.

{\bf CelebA}. The multiple attribute translation with different intensities, i.e., $\alpha$ = 0.25, 0.5, 0.75 and 1, are shown in Fig. \ref{CelebA_intensity}, where the face of a young woman is translated into various attributes by different methods. When the value of the attribute intensity is continuously changed, the images translated by StarGAN, AttGAN, and STGAN merely adapt the wrinkle translations. For the images translated by SwitchGAN and Gated SwitchGAN, they represent different attribute characteristics when different intensity values are imposed. For example, when the face of a young female is translated to an old man, the higher the intensity we impose, the shorter the hair and the higher the hair line that appear in the output. The images translated by Gated SwitchGAN are smoothly controlled by the intensity value and considerably fewer artefacts can be observed. While the intensities of translation for different attributes are set the same as the values in previous examples, we further show the translation results of two example faces in Figure 9, when the intensities of the attribute translation are set to four different values, i.e., 0, 0.25, 0.5 and 1. As shown in the figure, our Gated SwitchGAN can actually control the intensity of different attributes during face translation. Taking the face of the first lady as an example, when we are translating the female face to the male face, with an increasing value of $\alpha_{1}$, such that the face appears more similar to a male face (from left to right in the first row), we can actually decrease the value of $\alpha_{2}$ to reduce the amount of moustache. Similar results can also be achieved for other attributes such as hair colour and age.

\subsection{Ablation study}
We conduct ablation experiments to validate the effectiveness of the three components in our method: (1) self-reconstruction loss $\bm{L_{self}}$, (2) gate module, and (3) conditional feature-matching loss $\bm{L_{cfm}}$.

{\bf Self-reconstruction loss.} We first improve the SwitchGAN by adding self-reconstruction loss defined in Equation \ref{func12} to the training process, which preserves the global structure and unmodiﬁed attributes of the input face. As shown in Table \ref{ablation_study}, $\bm{L_{self}}$ can reasonably improve the FID/accuracy of SwitchGAN. We then denote the SwitchGAN with $\bm{L_{self}}$ as the {\bf baseline} for the following analysis of our proposed components.

{\bf Gate module.} We seek to validate that our gate module can effectively control the intensity of attribute translation and meanwhile improve the performance of image translation. We visually and quantitatively compare the performance of the {\bf baseline} and {\bf baseline+Gate} in Fig. \ref{ablation_study_for_gate}. As shown in the figure, the example images of Morph and RaFD translated by the {\bf baseline} remain stable for different intensities, and the faces are generally blurry, due to the loss of details. In contrast, our approach with the gate module ({\bf baseline+Gate}) can well control the intensity of age, expression, hair/gender and age/gender translations and generates considerably fewer artefacts than the {\bf baseline}.

{\bf Conditional feature matching loss.} We now test the usefulness of the conditional feature matching loss $\bm{L_{cfm}}$, by combining it with gate modules and comparing with different settings. Fig. \ref{ablation_study_for_cfm} shows the performance of our module with and without integrating $\bm{L_{cfm}}$. By exploiting domain information to constrain the generator, {\bf baseline+Gate+$\bm{L_{cfm}}$} effectively enhances the attribute manipulation ability and reduces artefacts in the generated images. As shown in Fig. \ref{ablation_study_for_cfm} (a), the face less than 20 years old translated with $\bm{L_{cfm}}$ regrows more black hair and presents less moustache to make the face look young. In contrast, the face older than 50 translated with $\bm{L_{cfm}}$ presents whiter beard/hairs and less hair. Similarly, as shown in Fig. \ref{ablation_study_for_cfm} (b) and Fig. \ref{ablation_study_for_cfm} (c), the quality of faces translated with $\bm{L_{cfm}}$ for different expressions and attributes are generally higher and look more realistic than that translated without $\bm{L_{cfm}}$. 

{\bf Quantitative analysis of different components.} Table \ref{ablation_study} lists the FID and Accuracy (Acc) of face images translated with different components, together with that of the {\bf baseline}. The general feature-matching loss defined in Equation \ref{func15}, $\bm{L_{fm}}$, is also included for comparison. As shown in the table, the gate module can generally improve both the accuracy and FID of {\bf baseline}. Taking RaFD as an example, the gate module improves the FID/accuracy of the {\bf baseline} from 27.45/95.23\% to 19.14/97.22\%. However, the inclusion of the $\bm{L_{fm}}$ loss function actually deteriorates the accuracy and increases the FID score of the baseline. It turns out that the combination of the gate and $\bm{L_{cfm}}$ achieves the best FID/accuracy among different settings.

\section{Conclusion}
In this paper, a generative adversarial network, named SwitchGAN, is proposed to achieve delicate multi-domain facial image translation among different tasks. A feature-switching operation and the conditional switch modules based on the new operation are proposed to introduce domain distribution matching for our SwitchGAN. As demonstrated in the paper, compared with the StarGAN, AttGAN and STGAN, SwitchGAN shows more remarkable performance for different tasks, such as age progression and regression, facial expression translation, and multiple facial attribute translation. We significantly extend our previous work by incorporating a gate module into the SwitchGAN. The extended SwitchGAN, named Gated SwitchGAN, achieves a markedly superior performance with the capability of controlling the intensity of face translation. The Gated SwitchGAN generator further shows a great potential for attribute intensity control. Much more intensive experiments have been conducted and popular image quality indices such as the FID score have also been applied to evaluate the quality of synthesized face images. We visually and quantitatively demonstrated the superiority of the SwitchGAN and Gated SwitchGAN models. Our future work may try to extend our model to other tasks such as style transfer by integrating specifically designed modules into the SwitchGAN.

\section*{Acknowledgement}
The work is supported by the National Natural Science Foundation of China (Grant Nos. 61672357 and U1713214), and the Science and Technology Project of Guangdong Province (Grant No. 2018A050501014).

\bibliographystyle{IEEEtran}
\bibliography{Reference_list}

    \begin{IEEEbiography}[{\includegraphics[width=1in,height=1.25in,clip,keepaspectratio]{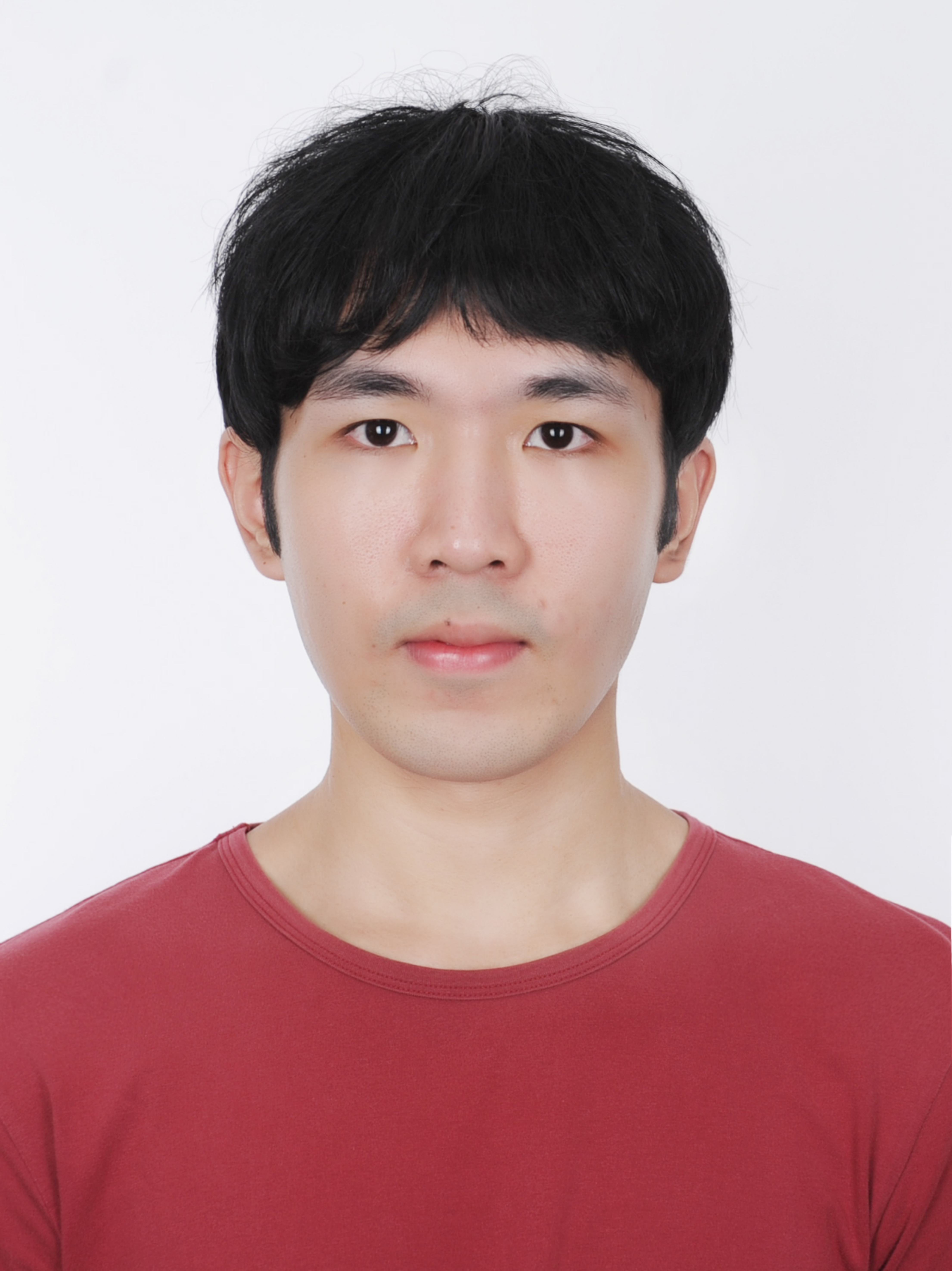}}]{Xiaokang Zhang}
	is currently a M.Sc. student majoring pattern recognition at School of Computer Science \& Software Engineering, Shenzhen University, China. He received the B.Sc. degree at School of Computer Science \& Software Engineering from Shenzhen University in 2018. His research interest includes Text-to-Image generation and face editing.
	\end{IEEEbiography}

	\begin{IEEEbiography}[{\includegraphics[width=1in,height=1.25in,clip,keepaspectratio]{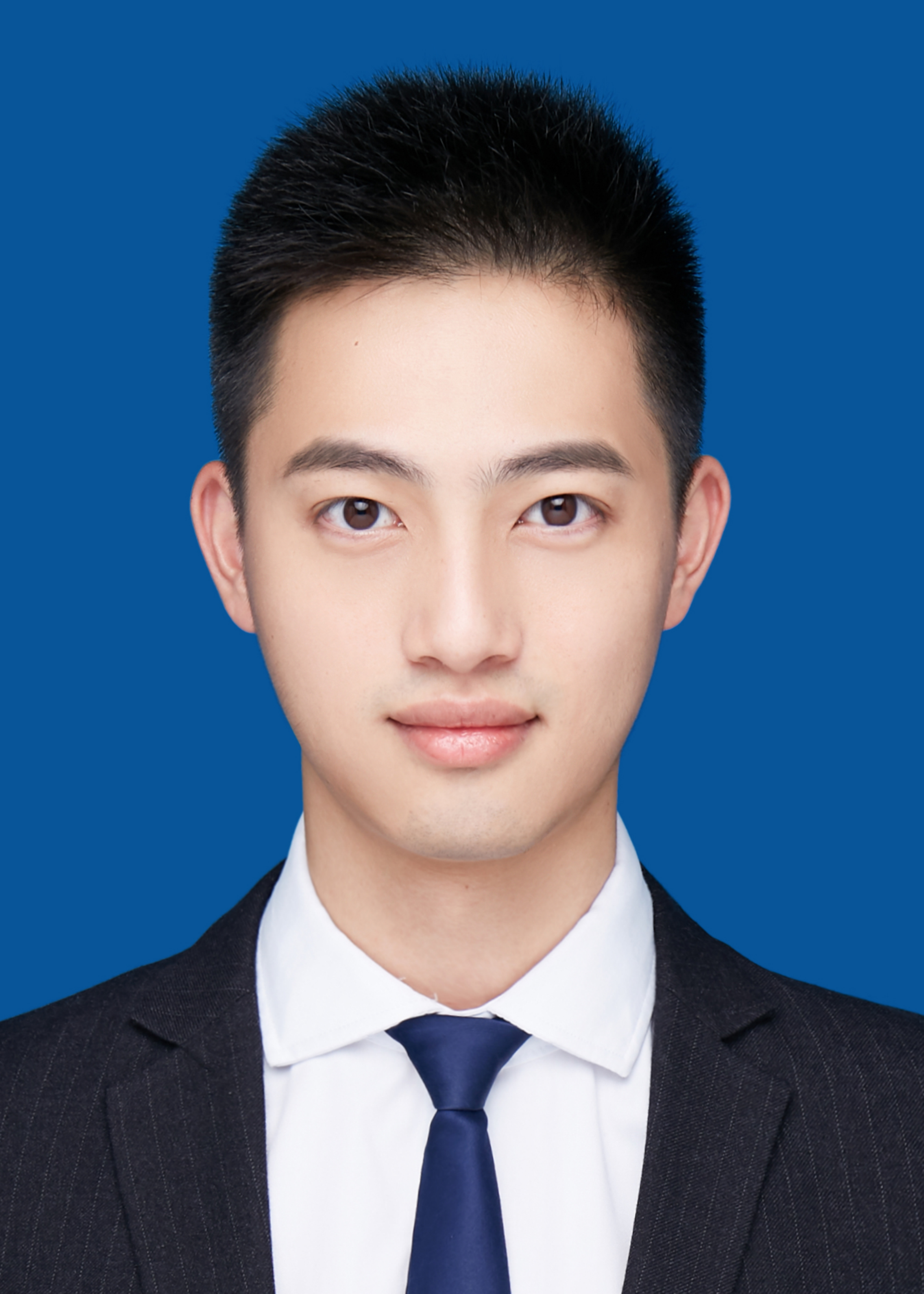}}]{Yuanlue Zhu}
		is currently a computer vision engineer at Bytedance. He received M.Sc. degree at School of Computer Science \& Software Engineering from Shenzhen University, China in 2020.  He received the B.Sc. degree of Applied Physics in South China University of Technology in 2017. His research interest includes face editing and image segmentation.
	\end{IEEEbiography}

	\begin{IEEEbiography}[{\includegraphics[width=1in,height=1.25in,clip,keepaspectratio]{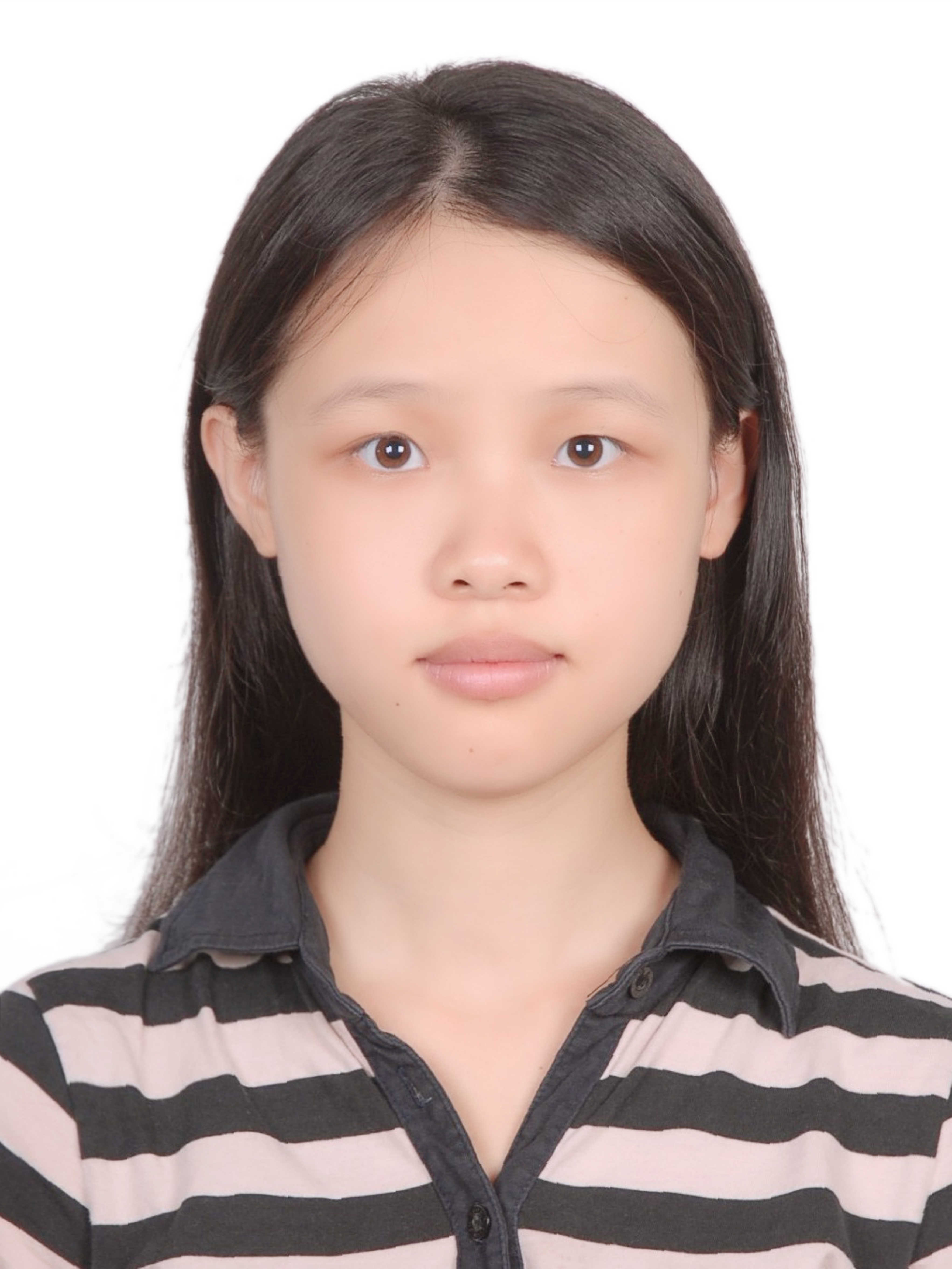}}]{Wenting Chen}
		is currently a Ph.D. student at The City University of Hong Kong, advised by Prof. Yixuan Yuan. Before that, she received her M.Sc. degree at School of Computer Science \& Software Engineering, Shenzhen University, China in 2020. She received the B.Sc. degree in educational technology and software engineering (double major) from Shenzhen University in 2017. Her research interest includes medical image processing and face editing.
	\end{IEEEbiography}
	
	\begin{IEEEbiography}[{\includegraphics[width=1in,height=1.30in,clip,keepaspectratio]{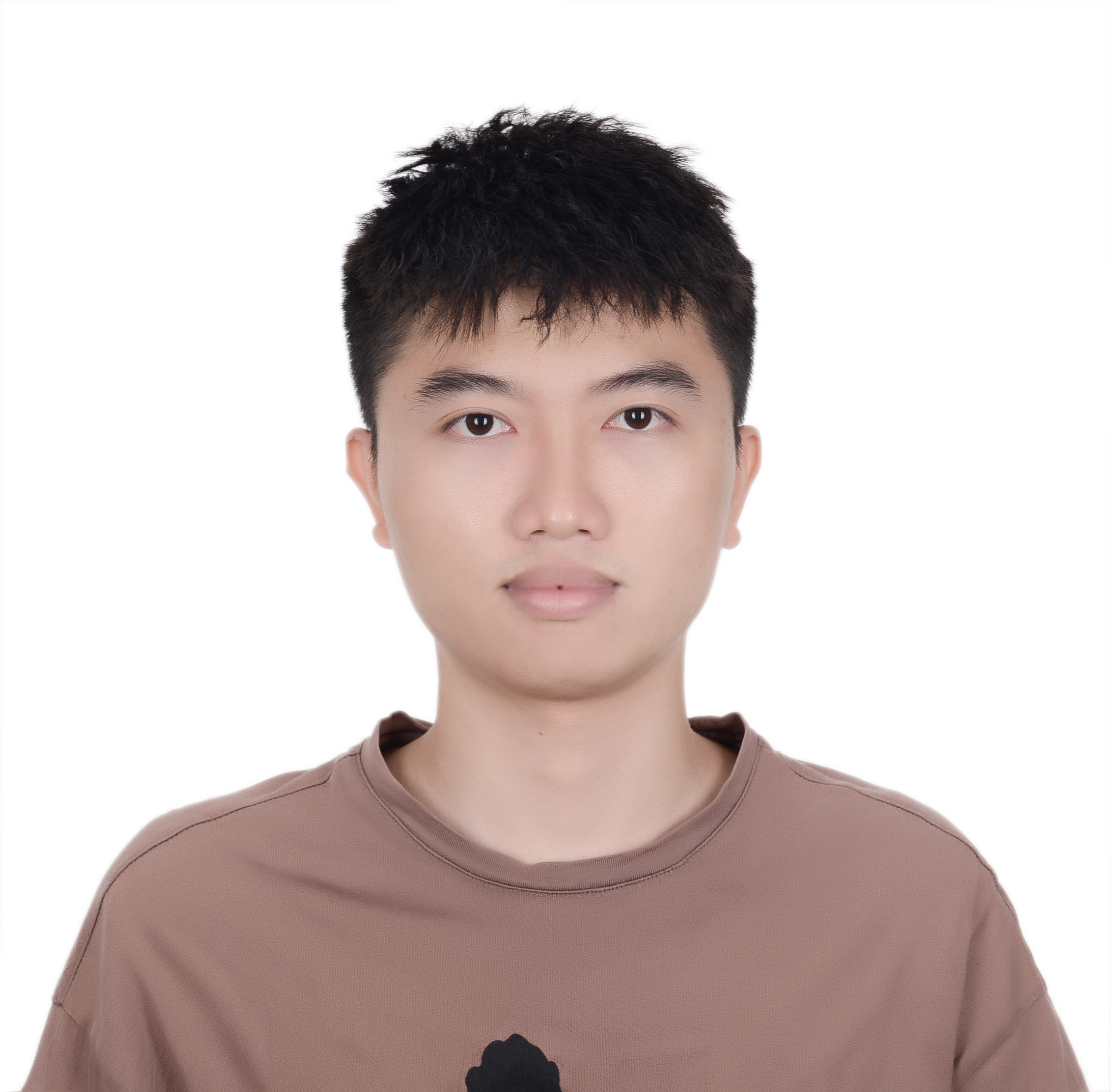}}]{Wenshuang Liu}
		is currently a M.Sc. student majoring pattern recognition at School of Computer Science \& Software Engineering, Shenzhen University, China. He received the B.Sc. degree at School of software engineering from South China Normal University in 2018. His research interest includes face recognition and editing.
	\end{IEEEbiography}

	\begin{IEEEbiography}[{\includegraphics[width=1in,height=1.25in,clip,keepaspectratio]{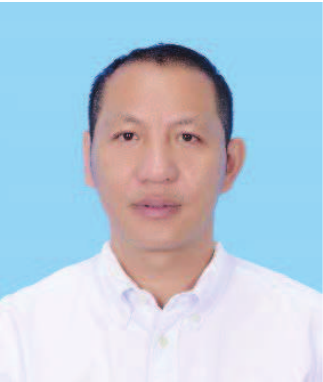}}]{Linlin Shen}
		is currently a Pengcheng Scholar Distinguished Professor at School of Computer Science and Software Engineering, Shenzhen University, Shenzhen, China. He is also a Honorary professor at School of Computer Science, University of Nottingham, UK. He serves as the director of Computer Vision Institute, AI Aided Medical Image Analysis \& Diagnosis Research Center and China-UK joint research lab for visual information processing. Prof. Shen received the BSc and MEng degrees from Shanghai Jiaotong University, Shanghai, China, and the Ph.D. degree from the University of Nottingham, Nottingham, U.K. He was a Research Fellow with the University of Nottingham, working on MRI brain image processing.
		
		His research interests include deep learning, facial recognition, analysis/synthesis and medical image processing. Prof. Shen is listed as the Most Cited Chinese Researchers by Elsevier. He received the Most Cited Paper Award from the journal of Image and Vision Computing. His cell classification algorithms were the winners of the International Contest on Pattern Recognition Techniques for Indirect Immunofluorescence Images held by ICIP 2013 and ICPR 2016.
	\end{IEEEbiography}

\ifCLASSOPTIONcaptionsoff
  \newpage
\fi

\end{document}